\pdfoutput=1
\documentclass[sigconf,authorversion]{acmart}

\usepackage[american]{babel}
\usepackage[hypcap]{subcaption}
\usepackage{anyfontsize,csquotes}
\usepackage{tikzexternal}
\tikzsetexternalprefix{figures/external/}
\newcommand{\norm}[1]{\left\lVert#1\right\rVert}

\definecolor{c1}{HTML}{EE7733} 
\definecolor{c2}{HTML}{0077BB} 
\definecolor{c3}{HTML}{009988} 
\definecolor{c4}{HTML}{EE3377} 
\definecolor{c5}{HTML}{33BBEE} 
\definecolor{gr}{HTML}{BBBBBB} 
\definecolor{acc}{HTML}{CC3311} 

\title[$(k, \ell)$-Medians Clustering of Trajectories Using Continuous Dynamic Time Warping]{\texorpdfstring{$(k, \ell)$}{(k, ℓ)}-Medians Clustering of Trajectories Using\texorpdfstring{\\}{ }Continuous Dynamic Time Warping}

\begin{teaserfigure}
\centering
\includegraphics[width=.223\textwidth]{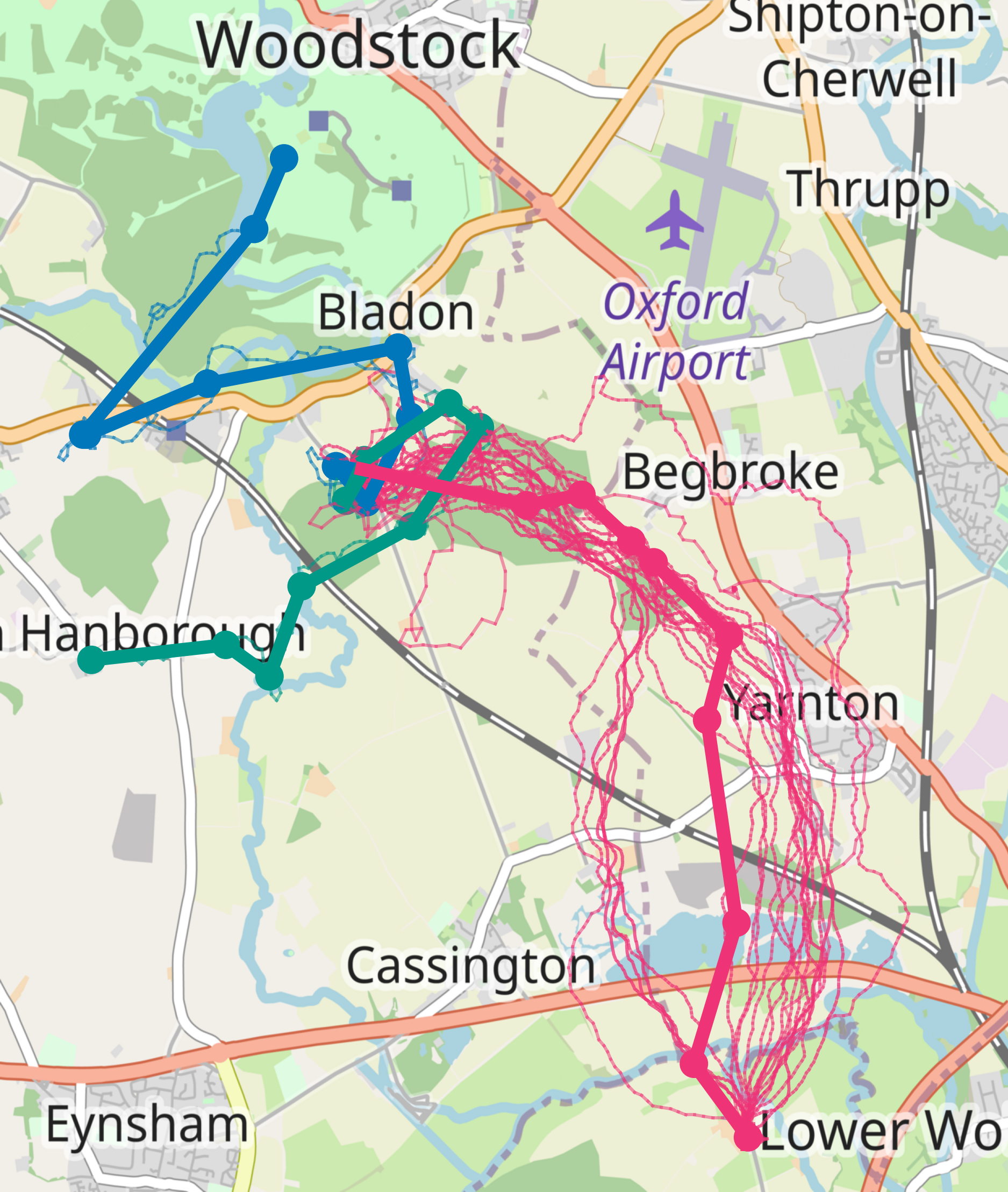}\hfil
\includegraphics[width=.223\textwidth]{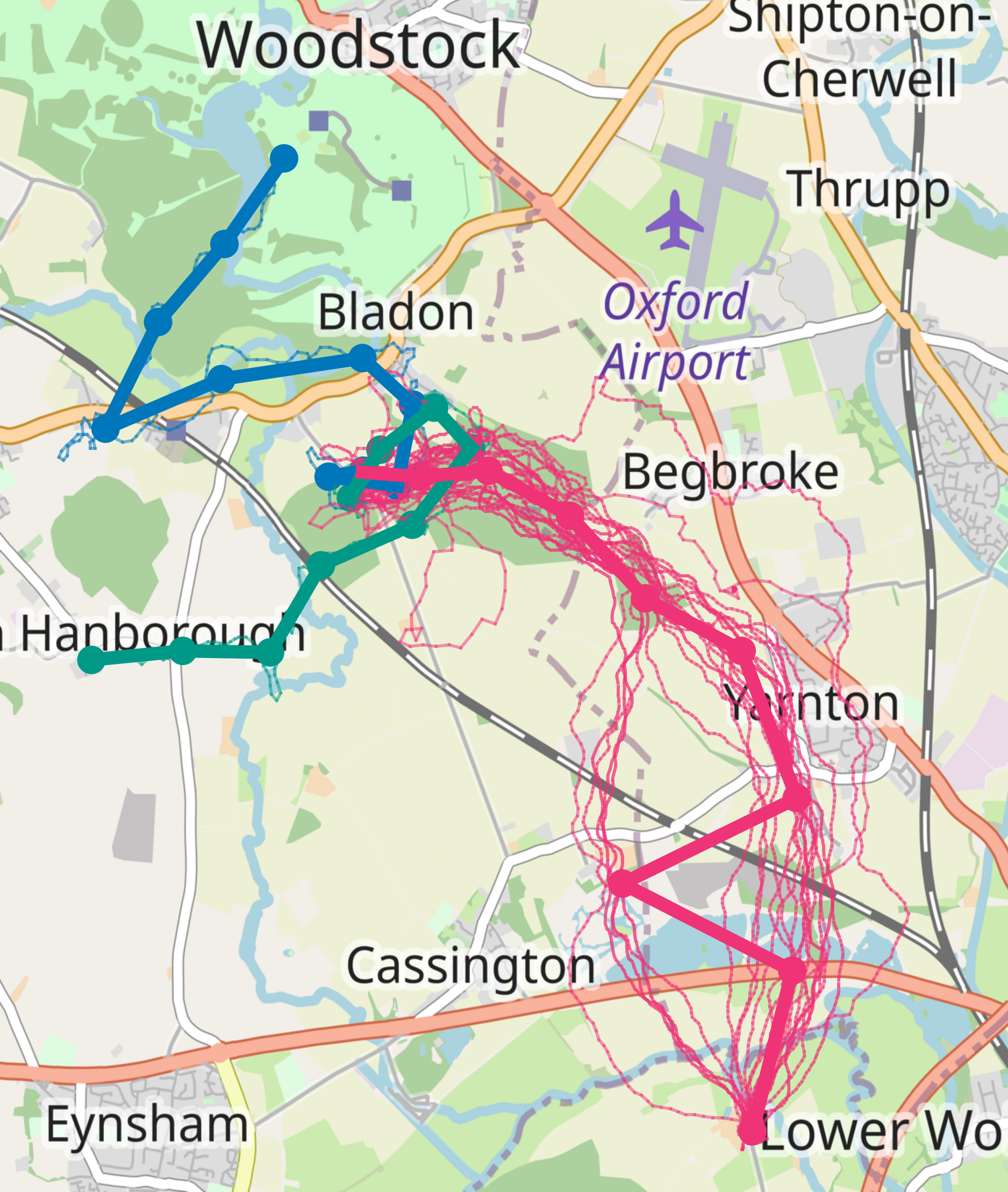}\hfil
\includegraphics[width=.223\textwidth]{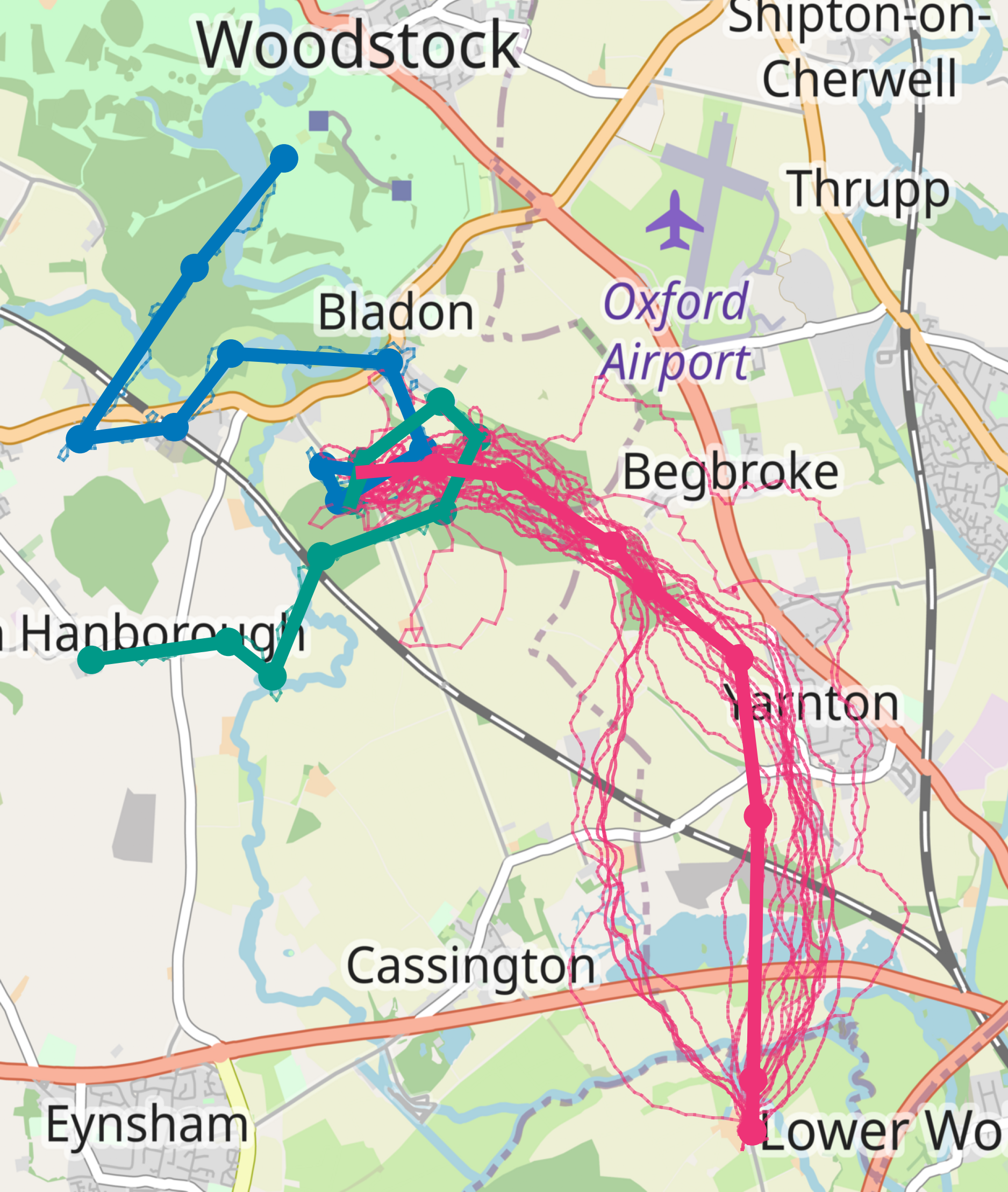}
\Description{Three subfigures with pigeon trajectories on the map.
The trajectories are clustered into three clusters; each cluster has a low-complexity center.
The Fr\'echet distance-based approach and the dynamic time warping-based approach exhibit some
artifacts in the cluster centers.
Our new approach looks most natural of the three.}
\caption{Clustering using the Fr\'echet distance (left), dynamic time warping (middle), and our new
approach called continuous dynamic time warping (right), where the color denotes the clusters and
the bold trajectories are the cluster centers.
While the Fr\'echet distance shows a strong influence of outliers and dynamic time warping shows
discretization issues, clustering via continuous dynamic time warping gives arguably the most
natural results.
Map data \textcopyright\ OpenStreetMap contributors.~\protect\cite{openstreetmap}}
\label{fig:pigeon_artifacts}
\end{teaserfigure}

\author{Milutin Brankovic}
\authornote{Authors listed in alphabetical order. All authors contributed equally to this research.}
\affiliation{
    \institution{University of Sydney}
    \streetaddress{1 Cleveland Street}
    \city{Darlington}
    \state{NSW}
    \postcode{2008}
    \country{Australia}
}
\email{mbra7655@uni.sydney.edu.au}
\author{Kevin Buchin}
\authornotemark[1]
\orcid{0000-0002-3022-7877}
\affiliation{
    \institution{TU Eindhoven}
    \streetaddress{Groene Loper 5}
    \postcode{5612\,AE}
    \city{Eindhoven}
    \country{Netherlands}
}
\email{k.a.buchin@tue.nl}
\author{Koen Klaren}
\authornotemark[1]
\affiliation{
    \institution{TU Eindhoven}
    \streetaddress{Groene Loper 5}
    \postcode{5612\,AE}
    \city{Eindhoven}
    \country{Netherlands}
}
\email{k.r.klaren@student.tue.nl}
\author{Andr\'e Nusser}
\authornotemark[1]
\affiliation{
    \institution{Max Planck Institute for Informatics}
    \institution{Graduate School of Computer Science}
    \streetaddress{Campus E1 4, Saarland Informatics Campus}
    \postcode{66123}
    \city{Saarbr\"ucken}
    \country{Germany}
}
\email{anusser@mpi-inf.mpg.de}
\author{Aleksandr Popov}
\authornotemark[1]
\orcid{0000-0002-0158-1746}
\affiliation{
    \institution{TU Eindhoven}
    \streetaddress{Groene Loper 5}
    \postcode{5612\,AE}
    \city{Eindhoven}
    \country{Netherlands}
}
\email{a.popov@tue.nl}
\author{Sampson Wong}
\authornotemark[1]
\affiliation{
    \institution{University of Sydney}
    \streetaddress{1 Cleveland Street}
    \city{Darlington}
    \state{NSW}
    \postcode{2008}
    \country{Australia}
}
\email{swon7907@uni.sydney.edu.au}


\copyrightyear{2020}
\acmYear{2020}
\setcopyright{acmlicensed}
\acmConference[SIGSPATIAL '20]{28th International Conference on Advances in Geographic Information Systems}{November 3--6, 2020}{Seattle, WA, USA}
\acmBooktitle{28th International Conference on Advances in Geographic Information Systems (SIGSPATIAL '20), November 3--6, 2020, Seattle, WA, USA}
\acmPrice{15.00}
\acmDOI{10.1145/3397536.3422245}
\acmISBN{978-1-4503-8019-5/20/11}


\begin{document}
\begin{abstract}
Due to the massively increasing amount of available geospatial data and the need to present it in an
understandable way, clustering this data is more important than ever.
As clusters might contain a large number of objects, having a representative for each cluster
significantly facilitates understanding a clustering.
Clustering methods relying on such representatives are called \emph{center-based.}
In this work we consider the problem of center-based clustering of trajectories.

In this setting, the representative of a cluster is again a trajectory.
To obtain a compact representation of the clusters and to avoid overfitting, we restrict the
complexity of the representative trajectories by a parameter $\ell$.
This restriction, however, makes discrete distance measures like dynamic time warping (DTW) less
suited.

There is recent work on center-based clustering of trajectories with a continuous distance measure,
namely, the Fr\'echet distance.
While the Fr\'echet distance allows for restriction of the center complexity, it can also be
sensitive to outliers, whereas averaging-type distance measures, like DTW, are less so.
To obtain a trajectory clustering algorithm that allows restricting center complexity and is more
robust to outliers, we propose the usage of a continuous version of DTW as distance measure, which
we call \emph{continuous dynamic time warping (CDTW).}
Our contribution is twofold:
\begin{enumerate}
    \item To combat the lack of practical algorithms for CDTW, we develop an approximation algorithm
    that computes it.
    \item We develop the first clustering algorithm under this distance measure and show a practical
    way to compute a center from a set of trajectories and subsequently iteratively improve it.
\end{enumerate}
To obtain insights into the results of clustering under CDTW on practical data, we conduct extensive
experiments.
\end{abstract}

\maketitle

\section{Introduction} \label{sec:introduction}
As smartphones have become a ubiquitous part of our daily lives, access to GPS technology has become
more convenient than ever.
In general, the prevalence of tracking technologies have led to an abundance of GPS data.
For example, researchers can track animals in large quantities to help study their behavior,
enabling new scientific insights in the field of behavioral biology.

With the blessing of easily available GPS data comes the burden of analyzing large quantities of
data, in particular, the need for suitable methods to extract relevant information from the
collected data.
A classical approach to make large data sets understandable is to cluster the data into different
groups.
While this makes it easy to identify different groups in the data, it may still be simply too much
information to digest or to visualize.
Identifying a good representative for each group helps condense the information even further.
Clustering methods that rely on such representatives---also called cluster centers---are referred
to as \emph{center-based clustering.}

Let a trajectory be represented by a polygonal chain in Euclidean space.
While point sets in Euclidean space have a canonical distance measure (the $L_2$-norm), for
trajectories there is no such clear choice.
Thus, trajectory clustering has been considered under several different distance measures.
Two popular distance measures are Dynamic Time Warping (DTW) and the Fr\'echet distance, but each
have their own shortcomings.

\begin{figure}
    \centering
    \begin{minipage}[b]{0.49\linewidth}
        \includegraphics[width=\textwidth]{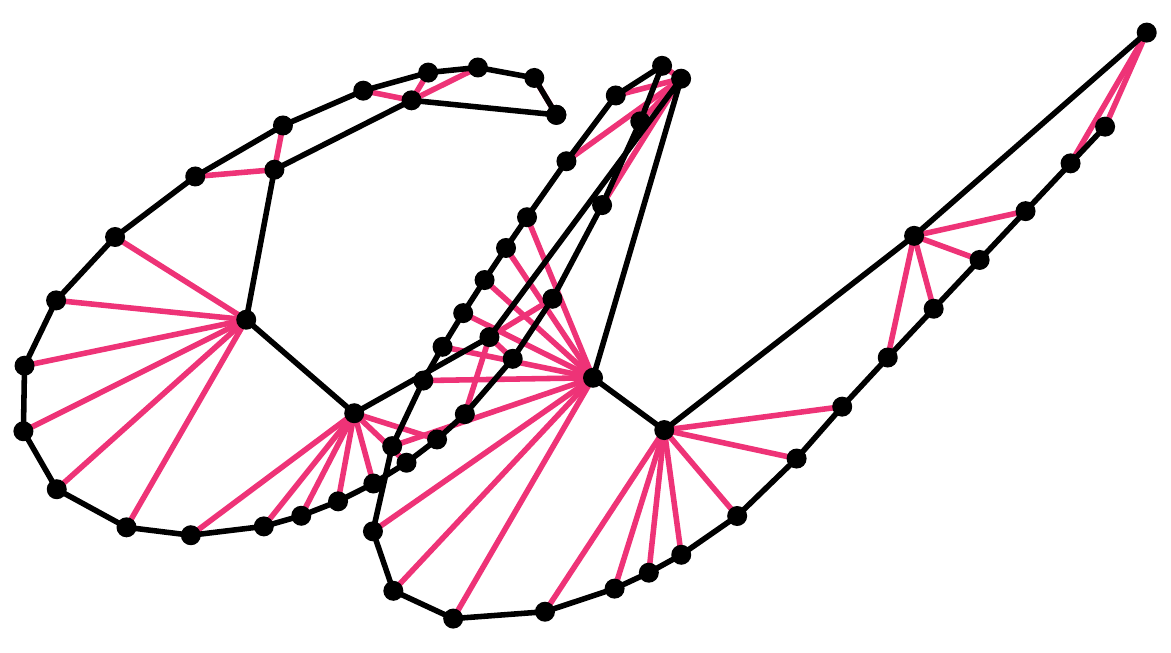}
        \subcaption{DTW.}
    \end{minipage}
    \begin{minipage}[b]{0.49\linewidth}
        \includegraphics[width=\textwidth]{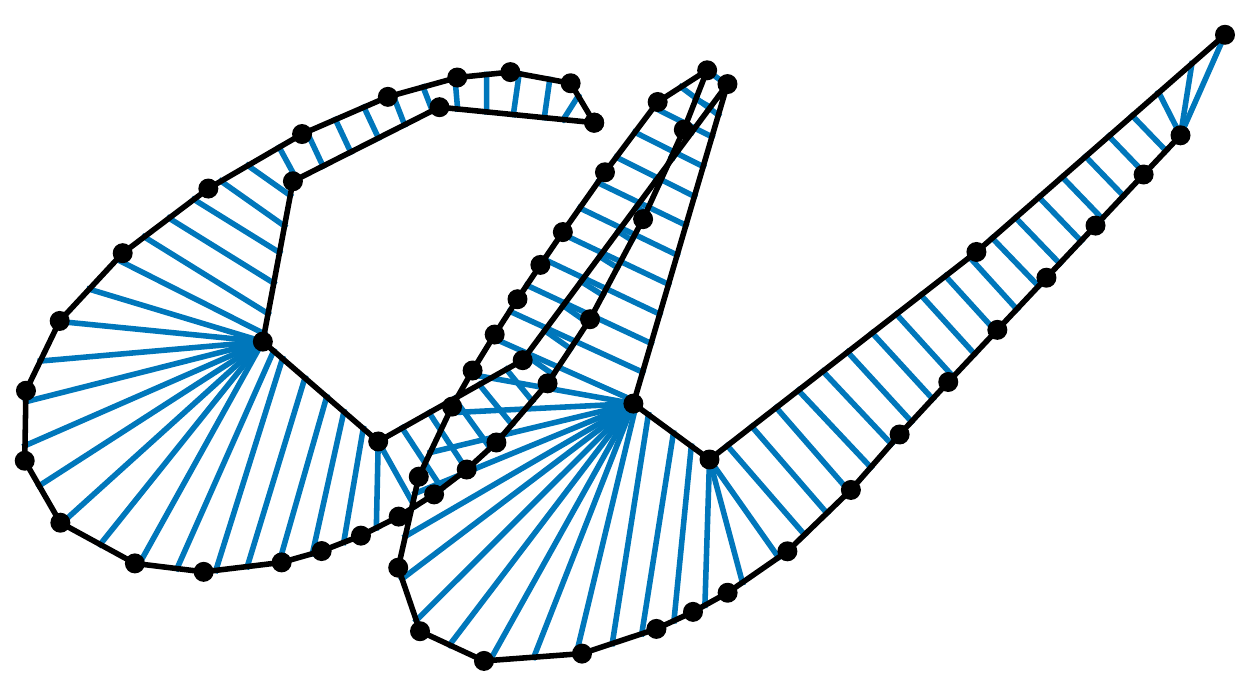}
        \subcaption{CDTW.}
    \end{minipage}
    \Description{Two trajectories and the corresponding matchings for DTW and CDTW.
    The DTW matching looks very unbalanced due to the sampling rate on the two trajectories, unlike
    the CDTW matching.}
    \caption{An example of discrete and continuous alignments of points along trajectories of
    differing complexities by DTW and CDTW, respectively.}
    \label{fig:warpings}
\end{figure}

\begin{figure}
    \centering
    \begin{minipage}[b]{0.49\linewidth}
        \includegraphics[width=\textwidth]{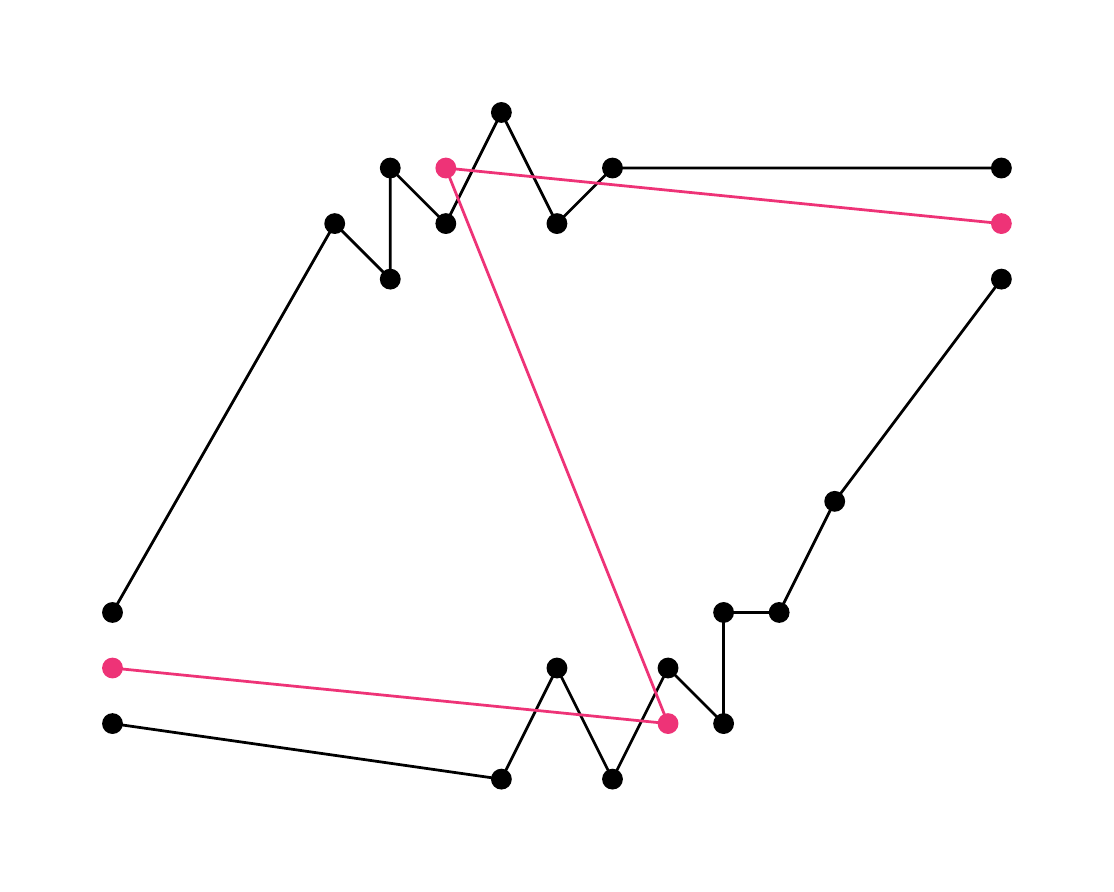}
        \subcaption{DTW cluster center.}
    \end{minipage}
    \begin{minipage}[b]{0.49\linewidth}
        \includegraphics[width=\textwidth]{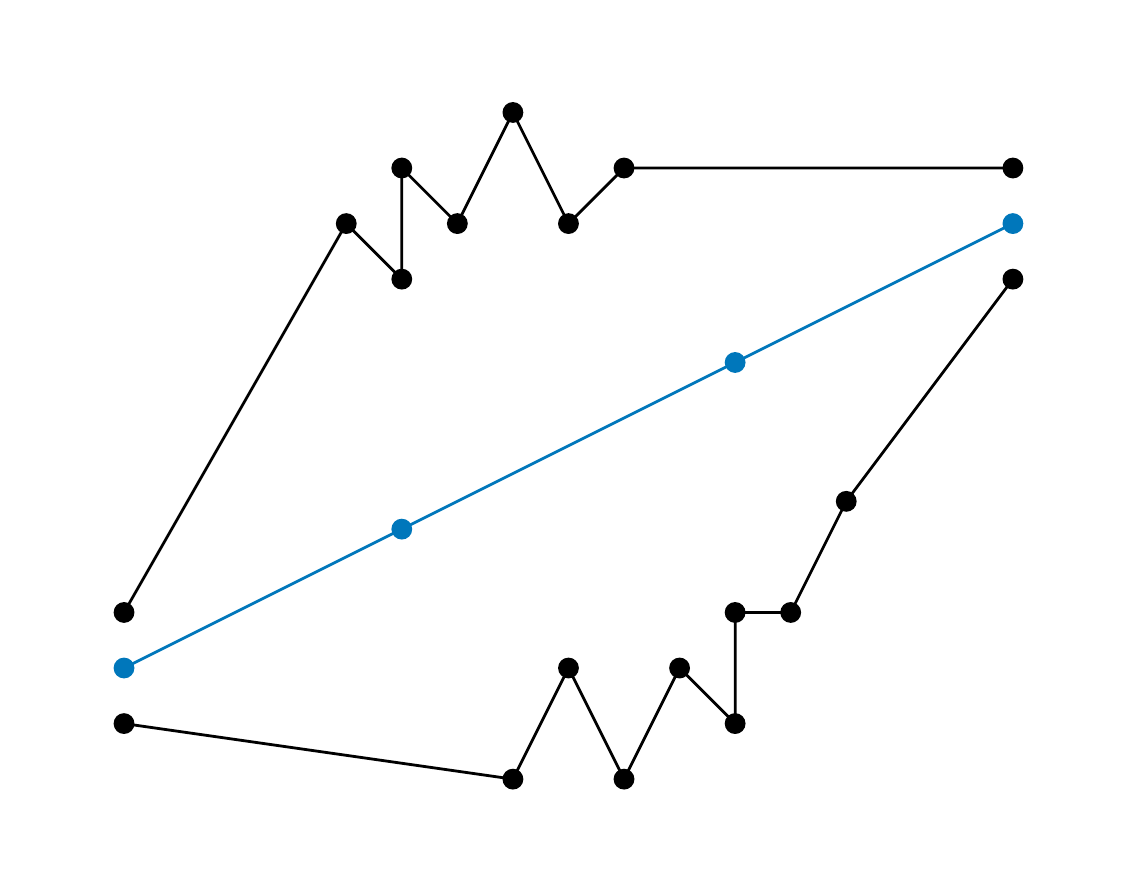}
        \subcaption{CDTW cluster center.}
    \end{minipage}
    \Description{Two trajectories and the corresponding low-complexity centers for DTW and CDTW.
    CDTW center goes smoothly between the trajectories, the DTW center is drawn first to one, then
    to the other trajectory.}
    \caption{An example where the low-complexity DTW cluster center is inappropriate for
    high-complexity trajectories.}
    \label{fig:dtw_clustering_artifact}
\end{figure}

Dynamic Time Warping (DTW) matches the vertices from one trajectory to vertices on the other such
that the summed distance between matched vertices is minimized.
In DTW, only vertices of the trajectory are considered, and the edges between these vertices are
ignored.
As such, DTW is sensitive to the relative sampling rates of the trajectories and provides a poor
matching between a high-complexity trajectory and a low-complexity trajectory.
See Figure~\ref{fig:warpings} for an example of this issue.
For this reason, DTW is inappropriate for computing low-complexity cluster centers for
high-complexity trajectories, as shown in Figure~\ref{fig:dtw_clustering_artifact}.

\begin{figure}
    \centering
    \begin{minipage}[b]{0.49\linewidth}
        \includegraphics[width=\textwidth,trim={0 2cm 0 0},clip]{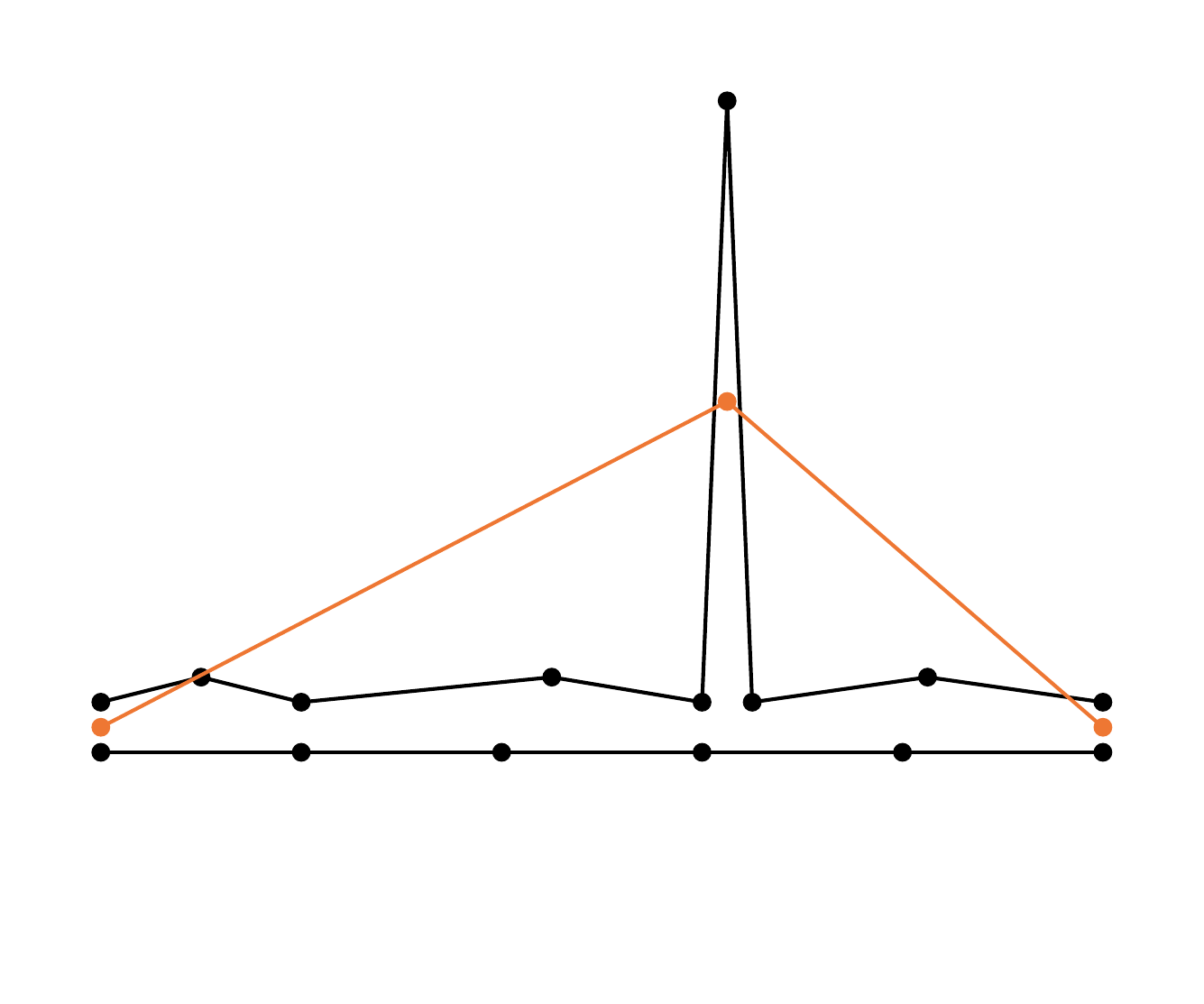}
        \subcaption{Fr\'echet cluster center.}
    \end{minipage}
    \begin{minipage}[b]{0.49\linewidth}
        \includegraphics[width=\textwidth,trim={0 2cm 0 0},clip]{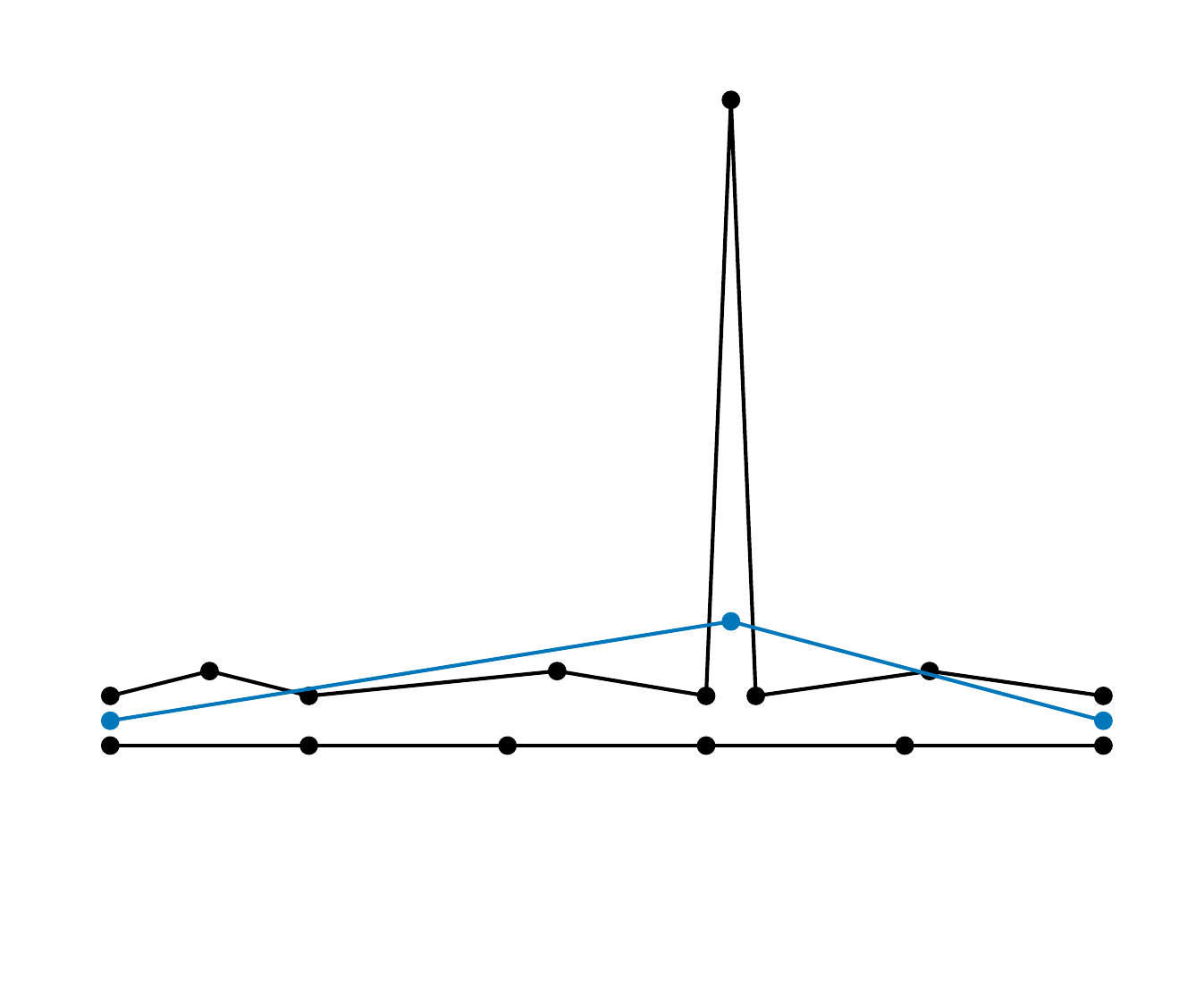}
        \subcaption{CDTW cluster center.}
    \end{minipage}
    \Description{Two trajectories, one of them with a sharp peak away from the other one in the
    middle, and the corresponding low-complexity centers for the Fr\'echet distance and CDTW.
    The CDTW center is only slightly perturbed at the sharp peak, unlike the Fr\'echet center.}
    \caption{An example where the quality of the Fr\'echet cluster center deteriorates in the
    presence of an outlier.}
    \label{fig:frechet_clustering_artifact}
\end{figure}

The Fr\'echet distance captures the minimal cost of a continuous deformation of one trajectory into
another, where the cost of deformation is the maximum distance between a point and its transformed
counterpart.
Recent work on center-based clustering of trajectories has used the Fr\'echet
distance~\cite{buchin2019,swat2020,driemel2016}.
The differences of DTW and the Fr\'echet distance are, first, that the Fr\'echet distance considers
both edges and vertices of the trajectory, and second, that the Fr\'echet distance is a bottleneck
measure (i.e., its cost is a maximum) whereas DTW is a summed measure.
As a result of the Fr\'echet distance being a bottleneck measure, a shortcoming of using the
Fr\'echet distance to compute cluster centers is that it may be sensitive to outliers in the
trajectory data.
See Figure~\ref{fig:frechet_clustering_artifact} for an example where this issue occurs.

To overcome the shortcomings of DTW and the Fr\'echet distance for center-based trajectory
clustering, we propose the usage of a continuous version of DTW.
We call this measure the \emph{continuous dynamic time warping distance (CDTW)} and formally define
it in Section~\ref{sec:cdtw}.
The advantages of computing cluster centers with CDTW are that it is appropriate for low-complexity
centers (see Figure~\ref{fig:dtw_clustering_artifact}) and is robust to outliers (see
Figure~\ref{fig:frechet_clustering_artifact}).
In Section~\ref{sec:experiments}, we provide empirical evidence in support of these claims.

The distance measure we call CDTW was originally introduced as the summed or average Fr\'echet
distance by Buchin~\cite[Ch.~6]{buchin2007}.
A different distance measure with the name CDTW was introduced by Munich and Perona~\cite{munich1999}.
This measure is continuous in the sense that a vertex of a trajectory may be aligned to any point on
the other trajectory, instead of only to another vertex.
It is, however, still based on discrete summation over the vertices in the same manner as DTW and is
therefore still prone to issues regarding sampling of the trajectories.
A general definition for a continuous distance measure for trajectories using path integrals with
the name CDTW was given by Efrat et al.~\cite{efrat2007curve}.
They also sketch an approximation algorithm for a special case of our definition of CDTW, but it is
based on numerical methods for solving eikonal equations, rendering it impractical.
A $(1 + \varepsilon)$-approximation algorithm for our definition of CDTW was introduced by
Maheshwari et al.~\cite{maheshwari2015}.
Its running time is upper-bounded by $O(\zeta^4n^4/\varepsilon^2)$, where $n$ is the complexity of
the trajectories and $\zeta$ is the maximal ratio between the lengths of any pair of segments from
the trajectories.
This dependency on the ratio between segments makes the algorithm unsuited for the problem of
$(k, \ell)$-center based clustering, where one trajectory consists of many short segments and the
other of a few long segments.
As far as we are aware, there are no exact algorithms known for computing the CDTW distance.

For point sets, clustering algorithms that are parametrized by the number of clusters, like
$k$-center or $k$-means clustering, are very common.
Trajectories, however, pose the problem that their complexity can vary.
Increasing the complexity of the centers also increases the adaptability of the centers to the data
at the cost of overfitting and complicated centers.
Therefore, recent work introduced the notion of $(k, \ell)$-clustering for
trajectories~\cite{buchin2019} (and there has been a previous closely related notion for time
series~\cite{driemel2016}), which is a center-based clustering where we require the clustering to
have $k$ clusters and each center can have complexity at most $\ell$.
More precisely, we consider the following two variants of $(k, \ell)$-clustering:
\begin{enumerate}
    \item\emph{$(k, \ell)$-center clustering:} Find $k$ centers of complexity at most $\ell$ that
    minimize the maximal distance that any trajectory has to its closest center.
    \item\emph{$(k, \ell)$-medians clustering:} Find $k$ centers of complexity at most $\ell$ that
    minimize the sum of distances of each trajectory to its closest center.
\end{enumerate}
See Section~\ref{sec:clustering_algs} for a formal definition and detailed discussion.
Interestingly, different notions of clustering seem more natural for different distance measures.
As the Fr\'echet distance and $(k,\ell)$-center clustering minimize the maximal distance, it seems
natural to combine them.
Similarly, as DTW / CDTW and $(k, \ell)$-medians minimize a sum / integral, it again seems natural
to combine them.

An approximation algorithm for $(k, \ell)$-center clustering using the continuous Fr\'echet distance
was given by Buchin et al.~\cite{buchin2019}.
This algorithm, with additional practical modifications, was later implemented and applied to
real-world data by Buchin et al.~\cite{buchin2019-sigspatial}.
For the discrete Fr\'echet distance, Buchin et al.~\cite{swat2020} designed approximation algorithms
for $(k, \ell)$-medians clustering and $(k, \ell)$-center clustering.
Finally, Petitjean et al.~\cite{petitjean} introduced an algorithm for computing low-complexity
centers for a cluster of trajectories using DTW distance.

\section{Contributions}
Our main contribution is a trajectory clustering implementation using the CDTW distance measure.
The advantages of this measure over existing measures are that it is appropriate for low-complexity
centers and robust to outliers.
To the best of our knowledge, our implementation is the first to cluster trajectories using CDTW.
The two main contributions to enable such an implementation are:
\begin{itemize}
\item As far as we are aware, there are no exact algorithms for CDTW, and the existing approximation
algorithms have a high dependency on both the complexity and the lengths of the input
trajectories~\cite{maheshwari2015}.
We overcome this obstacle by providing a practically efficient additive approximation algorithm for
computing the CDTW distance, and an efficient implementation of said algorithm.
See Section~\ref{sec:computing_cdtw} for details.
\item Again, as far as we are aware, there are no known algorithms for $(k, \ell)$-medians
clustering of trajectories under CDTW.
We overcome this obstacle by following the method of Buchin et al.~\cite{buchin2019}.
This involves computing an initial $(k, \ell)$-clustering by combining the Gonzalez algorithm or PAM
with a trajectory simplification algorithm, and then improving the cluster centers using methods
similar to those found in the practical work by Buchin et al.~\cite{buchin2019-sigspatial}.
We explore several methods for improving cluster centers, and perform extensive experiments to
compare these methods.
See Section~\ref{sec:clustering_algs} for details.
\end{itemize}
Our experimental results show that our method, which clusters trajectories under the CDTW measure,
outperforms similar methods which cluster trajectories under either the Fr\'echet distance or the
DTW measure.
The improvement of the CDTW measure over the DTW measure is particularly noticeable for
low-complexity cluster centers, and over the Fr\'echet distance in the presence of outliers.
See Section~\ref{sec:experiments} for details.

\section{Continuous Dynamic Time Warping}\label{sec:cdtw}
We represent trajectories as polygonal chains, i.e., series of connected line segments.
A trajectory $P$ is given by the sequence of its vertices $(p_1, \dots, p_n)$, where each
$p_i \in \mathbb{R}^d$ for some $d \ge 1$.
The DTW distance and the CDTW distance are summations over distances between aligned points along
the trajectories.
Among the different possibilities for distance measures between points, we consider the (squared)
Euclidean distance to be the most natural choice.
Both regular and squared Euclidean distances are commonly used in DTW computation.
Furthermore, methods for center computation, such as the least-squares method for linear regression
as well as $k$-means clustering, also use the squared Euclidean distance.
Thus, in the remainder we also focus on the squared Euclidean distance measure, i.e., the distance
between two points $p, q \in \mathbb{R}^d$ is defined as $d(p, q) = \norm{p - q}_2^2$.

A central definition for the DTW distance is the notion of a \emph{discrete warping path.}
Given a pair of trajectories $P = (p_1, \dots, p_n)$ and $Q = (q_1, \dots, q_m)$, define
$W = (w_1, \dots, w_k)$ with $w_i \in \{1, \dots, n\} \times \{1, \dots, m\}$.
This warping path aligns vertices in $P$ with vertices in $Q$, meaning that $w_k = (i, j)$
corresponds to a match of vertex $p_i \in P$ to vertex $q_j \in Q$.
The following restrictions are commonly imposed on discrete warping paths:
\begin{enumerate}
    \item The first and last vertex in $P$ must be matched to the first and last vertex in $Q$,
    respectively.
    \item Each vertex in $P$ must be matched to at least one vertex in $Q$, and vice versa.
    \item The alignment may not move backwards on the trajectories. More precisely, the indices in
    the warping path must be monotonically increasing, i.e., $i_{k - 1} \leq i_k$ and
    $j_{k - 1} \leq j_k$.
\end{enumerate}
Note that from~(2) and~(3) it follows that the warping path can only step to neighboring vertices.
For DTW, the cost of a warping path is the sum of distances between aligned vertices.
The DTW distance $\delta_D(P, Q)$ between a pair of trajectories is then the value of the
minimal-cost warping path:
\[\delta_D(P, Q) \coloneqq \min_W \sum_{(i, j) \in W} d(p_i, q_j)\,.\]

\begin{figure}
    \centering
    \includegraphics[width=\linewidth]{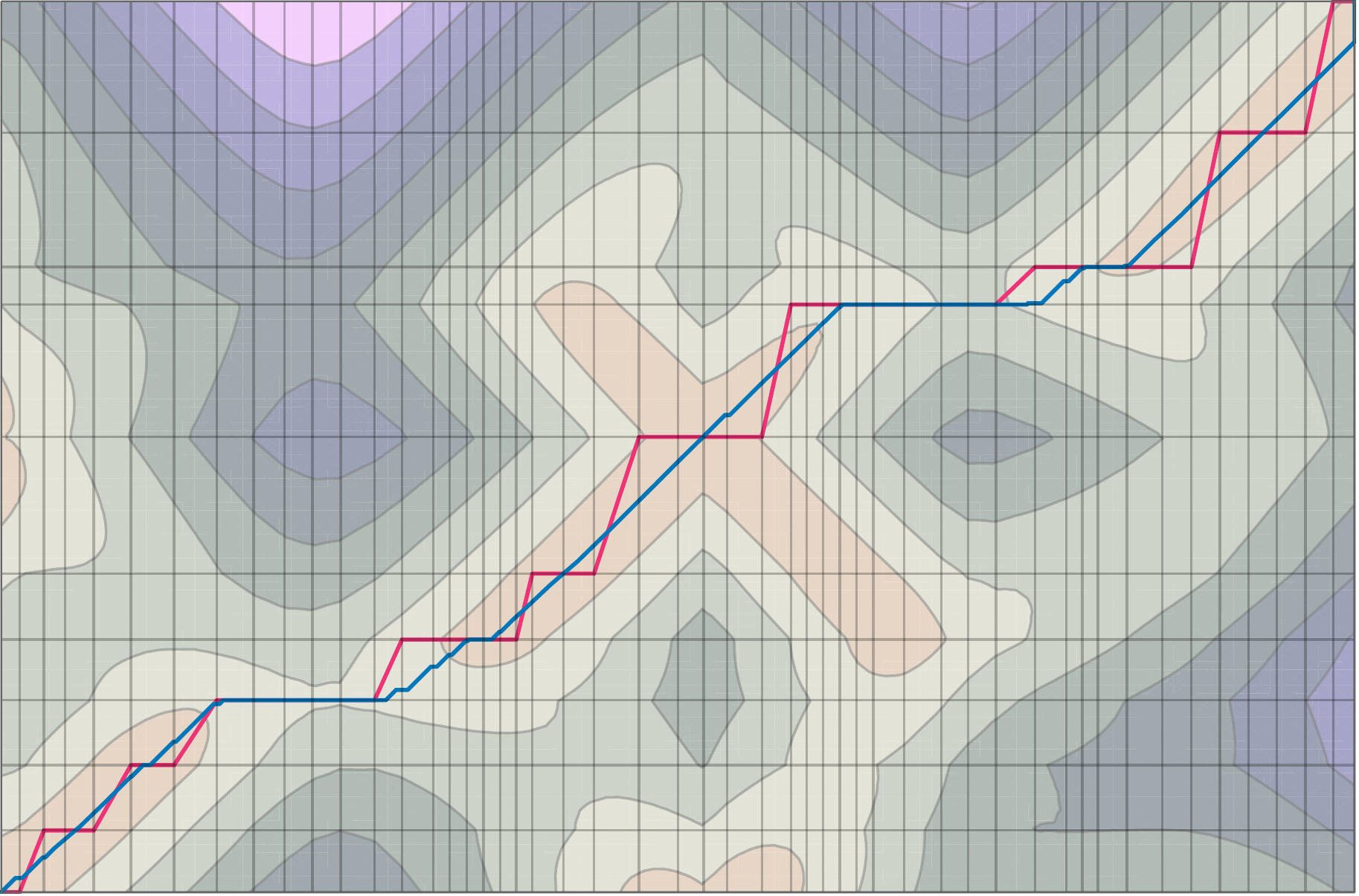}
    \Description{A figure showing the parameter space with a color map for matchings of higher and
    lower cost and two warping paths.
    The one for DTW goes from vertex to vertex, so horizontally, vertically, or diagonally.
    The one for CDTW can cross the grid lines in the parameter space at any point.}
    \caption{Warping paths of \textcolor{c4}{DTW} and \textcolor{c2}{CDTW} through parameter space
    corresponding to the trajectories in Figure~\ref{fig:warpings}.
    The grid lines correspond to the vertices of the trajectories.}
    \label{fig:parameter-space}
\end{figure}

Let $L(P)$ denote the total arc length of a trajectory $P$.
We define the parameter space $\mathbb{P} = [0, L(P)] \times [0, L(Q)]$ as an axis-aligned rectangle
representing all pairs of points in trajectories $P$ and $Q$.
Let $P(t)$ denote the point that lies arc length $t$ along $P$.
Then, each $(p, q) \in \mathbb{P}$ corresponds to a pair of points $(P(p), Q(q))$.
We define a height function $h\colon \mathbb{P} \to \mathbb{R}^+$ over the parameter space that
gives the distance between the corresponding points, i.e., $h((p, q)) \coloneqq d(P(p), Q(q))$.
Figure~\ref{fig:parameter-space} shows an example of the parameter space and height function for the
pair of trajectories in Figure~\ref{fig:warpings}.

A \emph{continuous warping path} $\pi \colon [0, 1] \to \mathbb{P}$ aligns points in $P$ to points
in $Q$, i.e., if $(p, q)$ is in the image of $\pi$, the point $P(p)$ is aligned to the point $Q(q)$.
We impose restrictions on continuous warping paths similar to those for its discrete counterpart.
\begin{enumerate}
    \item The first and last vertex in $P$ must be matched to the first and last vertex in $Q$,
    respectively.
    \item Every point in $P$ must be matched with at least one point in $Q$, and vice versa.
    \item The alignment may not move backwards over the trajectories. More precisely, both
    coordinates of $\pi$ must be monotonically increasing.
\end{enumerate}
It follows that $\pi$ is a continuous function that monotonically increases from $\pi(0) = (0, 0)$
to $\pi(1) = (L(P), L(Q))$.
For CDTW, the cost of a warping path is the integral over distances between aligned points.
Therefore, a natural definition is the path integral over $\pi$ through $h$:
\[\int_\pi h(r)\,\mathrm{d}r = \int_0^1 h(\pi(t)) \cdot \norm{\pi'(t)}\,\mathrm{d}t\,.\]
It may be desirable to use different norms for the speed of the path $\norm{\pi'(t)}$, so we
generalize the definition to support this.
The CDTW distance between a pair of trajectories using the $L_p$-norm in parameter space is the cost
of the continuous warping path with minimal cost:
\[\delta^p_{CD}(P,Q) \coloneqq \min_\pi \int_0^1 h(\pi(t)) \cdot \norm{\pi'(t)}_p\,\mathrm{d}t\,.\]
By using $p = 1$, the path length $\int_0^1 \norm{\pi'(t)}_1\,\mathrm{d}t = L(P) + L(Q)$ is the same
for all warping paths, as noted by Buchin~\cite{buchin2007}, thus allowing for normalization and
easier comparison.
In a similar setting, $p = \infty$ was used to place a speed limit on the warping
path~\cite{rote2014lexicographic}.
We identify $p = 1$ and $p = \infty$ as good choices for $p$, in particular due to their natural
meaning as the sum and the maximum of speeds along the trajectories.
In this work we use $p = 1$ and thus define $\delta_{CD}(P, Q) \coloneqq \delta^1_{CD}(P, Q)$.
An example of minimal-cost warping paths for both DTW and CDTW is shown in
Figure~\ref{fig:parameter-space}.

\section{Computing CDTW}\label{sec:computing_cdtw}
The task of computing the continuous dynamic time warping distance reduces to finding an optimal
warping path, i.e., a warping path for which the CDTW cost is minimized.
We first split the problem into smaller subproblems which can easily be solved exactly, and then
apply existing algorithms to combine the exact solutions into an approximation of an optimal warping
path.

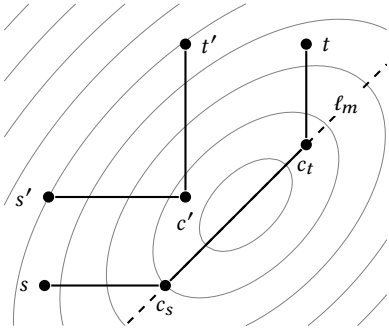
\begin{figure}
    \centering
    \tikzsetnextfilename{warping-path}
\begin{tikzpicture}
\def\a{0}
\def\b{0}
\def\l{-1}

\def\sx{-5}
\def\sy{-2}
\def\tx{1.5}
\def\ty{4}

\def\xmin{-6}
\def\xmax{3.5}
\def\ymin{-3}
\def\ymax{5}

\begin{axis}[myaxis,width=0.6\linewidth]

\ellipses

\addplot[ellm, variable=\t](t, \ellmy{t});
\node[linelabel] at \ellmcoord{\a+2.5} {$\ell_m$};


\node[point, label=left:{$s$}] (s) at (axis cs:\sx,\sy) {};
\node[point, label=below:{$c_s$}] (p1) at (axis cs:\ellmx{\sy},\sy) {};
\node[point, label=below:{$c_t$}] (p2) at (axis cs:\tx,\ellmy{\tx}) {};
\node[point, label=right:{$t$}] (t) at (axis cs:\tx,\ty) {};
\draw[mymatching] 
    (s) edge (p1)
    (p1) edge (p2)
    (p2) edge (t);

\node[point, label=left:{$s'$}] (s2) at (axis cs:\sx+0.1,\sy+2.2) {};
\node[point, label=below:{$c'$}] (c2) at (axis cs:\tx-3,\sy+2.2) {};
\node[point, label=right:{$t'$}] (t2) at (axis cs:\tx-3,\ty) {};
\draw[mymatching]
    (s2) edge (c2)
    (c2) edge (t2);

\end{axis}
\end{tikzpicture}
    \Description{Two paths within a cell, depending on the starting and ending location.
    One follows the cell border, then the line $l_m$, then another cell border; the other does not
    reach $l_m$, so just follows the two borders.}
    \caption{Two types of optimal warping paths through a single cell: $(s, c_s, c_t, t)$ via
    $\ell_m$, and $(s', c', t')$ missing $\ell_m$.}
    \label{fig:optimal-warping-path}
\end{figure}

As follows from the description in Section~\ref{sec:cdtw}, the segments of the trajectories induce
a grid of axis-aligned rectangular cells over the entire parameter space, such that each cell
uniquely corresponds to a pair of segments.
See Figure~\ref{fig:parameter-space} for an example of such a parameter space with its grid of
cells.
The height function over a single cell is therefore a function over linear interpolations of the
segments.
More precisely, we define the height function at the point $(p, q) \in \mathbb{P}$ in a parameter
space cell as
\[h((p,q)) \coloneqq (p - a)^2 + (q - b)^2 + 2\lambda (p - a) (q - b) + c\,,\]
where the values for $a, b \in \mathbb{R}$ are based on the offset of the cell's corresponding
segments along the infinite lines on which they lie.
Note that these values are fixed for each cell.
The value of $-1 \leq \lambda \leq 1$ is the dot product between the unit direction vectors of the
segments and is fixed for each cell as well.
The value of $c$ is the smallest distance between the infinite lines of the segments and is
therefore non-zero only in the degenerate case of parallel lines.
The \emph{level sets} of the height function $h$ form concentric ellipses with center $(a, b)$ and
eccentricity based on $\lambda$, possibly degenerating into parallel strips.
Let $\ell_m$ be the line through the point $(a, b)$ with slope~$1$, coinciding with the
$xy$-monotone axes of the ellipses.
Figure~\ref{fig:optimal-warping-path} shows a number of these level sets and the $\ell_m$ line.

Given a cell $C$ and points $s, t \in C$ with $s_x \leq t_x$ and $s_y \leq t_y$ and assuming without
loss of generality that $s$ and $t$ are the bottom left and top right corners of $C$, respectively,
we can compute the optimal warping path from $s$ to $t$ as shown by Maheshwari et
al.~\cite{maheshwari2015}:
\begin{itemize}
    \item If $\ell_m$ intersects the border of $C$, let $c_s$ and $c_t$ be the points of
    intersection closest to $s$ and $t$, respectively.
    Then the optimal warping path is $(s, c_s, c_t, t)$.
    \item Otherwise, let $c$ be the corner point closest to $\ell_m$.
    Then the optimal warping path is $(s, c, t)$.
\end{itemize}
An example for both cases is shown in Figure~\ref{fig:optimal-warping-path}.
Note that in both cases the optimal warping path consists of a constant number of linear pieces.
We integrate along each linear piece to compute the cost of the warping path from $s$ to $t$.

\begin{figure}
    \centering
    \tikzsetnextfilename{graph}
\begin{tikzpicture}
\def\a{0}
\def\b{0}
\def\l{-1}

\def\sx{-6}
\def\sy{-2}
\def\tx{2}
\def\ty{4}

\def\xmin{\sx}
\def\xmax{\tx}
\def\ymin{\sy}
\def\ymax{\ty}

\begin{axis}[myaxis,width=0.6\linewidth]

\ellipses

\addplot[mymatching, variable=\t](t, \ellmy{t});

\draw[draw=black] (axis cs:\sx,\sy) rectangle (axis cs:\tx,\ty);

\end{axis}

\def\w{5.07}
\def\h{3.8}

\def\xn{7}
\foreach \x in {0,...,\xn} {
    \node[point] (b\x) at (\x/\xn*\w,0) {};
    \node[point] (t\x) at (\x/\xn*\w,\h) {};
    \coordinate (vx\x) at (\x/\xn*\w,\x/\xn*\w - \w/2);
}

\def\yn{5}
\foreach \y in {0,...,\yn} {
    \node[point] (l\y) at (0,\y/\yn*\h) {};
    \node[point] (r\y) at (\w,\y/\yn*\h) {};
    \coordinate (vy\y) at (\y/\yn*\h + \w/2,\y/\yn*\h);
}

\foreach \x in {2,3} {
    \draw[mymatching,->] (b\x) -- (t\x);
}
\foreach \x in {4,...,\xn} {
    \draw[mymatching,->] (vx\x) -- (t\x);
}
\foreach \y in {0,...,3} {
    \draw[mymatching,->] (vy\y) -- (r\y);
}
\draw[mymatching,->] (vx7) -- (r4);
\draw[mymatching] (b2) -- (vy0);

\node[label=below:$s$] at (b2) {};

\draw[decoration={brace,mirror,raise=5pt},decorate] 
    (b0.center) -- node[below=6pt] {$\leq \Delta$} (b1.center);

\draw[decoration={brace,raise=5pt},decorate] 
    (l0.center) -- node[left=6pt] {$\leq \Delta$} (l1.center);
\end{tikzpicture}
    \Description{The cell with regularly sampled entry and exit points on the borders. Depending on
    the level sets and the choice of points, the path through the cell may differ.}
    \caption{Example placement of vertices along the borders of a cell, and an example of the edges
    originating from one such vertex~$s$ visualized by the optimal warping paths.}
    \label{fig:parameter-space-graph}
\end{figure}
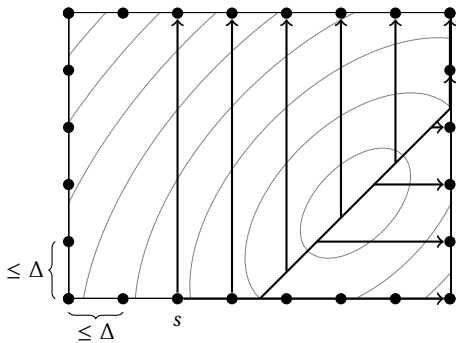

With this algorithm for computing the optimal warping path through single cells, it remains to find
the points where the optimal warping path crosses from one cell to another.
To this end, define a graph over the parameter space.
Take the corner points of the cells as the vertices of the graph.
To improve the accuracy of the warping path, we uniformly sample additional vertices (called Steiner
points) along the borders of each cell, such that the distance between two neighboring sampled
vertices is at most $\Delta$ in the original Euclidean space.
We add an edge between two vertices $s$ and $t$ if both vertices lie in the same cell and are
positioned such that $s_x \leq t_x$ and $s_y \leq t_y$.
Each edge is assigned a weight equal to the cost of the optimal warping path between its incident
vertices.
An example of vertex placement and edges is shown in Figure~\ref{fig:parameter-space-graph}.
The cost of the shortest path through this graph from $(0, 0)$ to $(L(P), L(Q))$ is an additive
approximation to the cost of the optimal warping path.

Any algorithm that finds the shortest path between two given vertices in an acyclic graph with
positive edge weights will be sufficient in finding the approximation of the optimal warping path.
In our implementation we use bidirectional Dijkstra's algorithm as it often needs to consider fewer
vertices than regular Dijkstra's algorithm.
This algorithm searches from both the start and goal vertex until the two search fronts meet with an
appropriate stopping condition.
In practice we do not compute the entire graph in advance; instead, for a given vertex, we compute
the adjacent vertices and incident edges on the fly as they are explored by the shortest path
algorithm.

\section{Clustering Algorithms}\label{sec:clustering_algs}
Before we present our clustering algorithms, we first formally define the two notions of clustering
that were presented in Section~\ref{sec:introduction}.
Let $\mathcal{C}$ be a set of trajectories and let $k$ and $\ell$ be positive integers.
Let $d$ be some distance measure for trajectories.
In a $(k, \ell)$-clustering problem, we seek to compute a set of $k$ center trajectories
$\mathcal{Q}$ of complexity at most $\ell$ which is optimal according to some cost function.
The two notions we consider are:
\begin{enumerate}
    \item In $(k, \ell)$-center clustering, the maximum distance from each trajectory in
    $\mathcal{C}$ to the nearest center in $\mathcal{Q}$ is minimized, i.e., the relevant cost
    function is
    \[\phi_{\infty}(Q)\coloneqq\max_{C\in \mathcal{C}}d(C, \mathcal{Q})\,.\]
    \item In $(k, \ell)$-medians clustering, the sum of distances from each trajectory in
    $\mathcal{C}$ to the nearest center in $\mathcal{Q}$ is minimized, i.e., the relevant cost
    function is
    \[\phi_{1}(Q)\coloneqq\sum_{C\in \mathcal{C}}d(C, \mathcal{Q})\,.\]
\end{enumerate}
For some variants we also consider $(k, \ell)$-medians clustering where we use the squared
$L_2$-norm as the underlying distance measure.
This can also be considered a means clustering.

Our high-level strategy for $(k, \ell)$-medians clustering is as follows.
We first compute an initial set of $k$ centers, each of them being an $\ell$-simplification of a
trajectory from the input set.
We then apply an iterative center improvement algorithm to obtain cluster centers that are not
restricted to the vertices of the input trajectories.

\subsection{Computing the Initial Clustering} \label{sec:init_cluster}
We use the \emph{partitioning around medoids (PAM)} method, also called $k$-medoids, proposed by
Kaufman and Rousseeuw~\citetext{\citealp{pam87}, \citealp[Ch.~2]{pam89}} and recently
improved by Schubert and Rousseeuw~\cite{pam2019}.
However, we adapt the method to the $(k, \ell)$-medians clustering setting.
The PAM method for $k$-medians clustering in general metric spaces works by first greedily choosing
an initial set of centers.
In this step, we compute the reduction in the sum of distances from trajectories to cluster centers
for each trajectory and then choose the one that gives the biggest reduction as a new center; we add
$k$ centers this way.
Next, the algorithm enters a local search phase where for each center $c_i$, we evaluate the
possible reduction in the sum of distances if we swap $c_i$ for a non-center $x$.
We perform the swap that gives the biggest difference and repeat the procedure iteratively until we
find a local optimum.
Adapting this approach to our setting is straightforward.
Whenever we consider a trajectory as a center, we compute the distances to its
$\ell$-simplification.

We also adapt the \emph{greedy algorithm of Gonzalez} to the setting of
$(k, \ell)$-clustering~\cite{gonzalez}.
The idea is simple: we start with a random input trajectory as the first center; then we iteratively
pick the trajectory which is furthest from the already picked centers until we get $k$ centers.
The only difference is that we pick the $\ell$-simplifications of the center trajectories, and
compute the distances accordingly.
Even though the Gonzalez algorithm optimizes for the $k$-center cost function, it is known to be a
$2n$-approximation for the $k$-medians problem, where $n$ is the number of input
points~\cite[Ch.~4]{har2011geometric}.
Therefore, we use the Gonzalez algorithm as a baseline for our initial clustering approaches.
We also consider an alternative clustering algorithm that first computes a $(k, \ell)$-clustering
using the Gonzalez algorithm and then applies the local search of PAM.

\subsection{Trajectory Simplification}\label{sec:simpl_desc}
To use the approach described in Section~\ref{sec:init_cluster}, we need to compute the
$\ell$-simplifications of the center trajectories when computing an
initial clustering.
The choice of simplification approach to use is somewhat independent of the clustering methods: we
need to choose an approach that allows us to restrict the maximum complexity of the simplified
trajectory to $\ell$ and gives good simplifications, both visually and in terms of CDTW.
As this is only used as the initial step before improving the centers, we choose the simpler
approach of vertex-restricted simplification, where the vertices of the simplified trajectory are a
subset of the vertices of the original trajectory.
We focus on two simplification methods: the dynamic programming approach suggested by Imai and
Iri~\cite{imai_iri} and the greedy approach by Agarwal et al.~\cite{agarwal_greedy}.

The greedy approach is very efficient, offers reasonable guarantees for the Fr\'echet distance, and
works well in practice.
The main idea is: given a threshold and a starting vertex of the trajectory $p_i$, go along the
trajectory until we reach a vertex $p_j$ such that the distance of choice between the segment
$p_ip_{j + 1}$ and the subtrajectory from $p_i$ to $p_{j + 1}$ exceeds the threshold.
Here $i$ and $j$ are indices of vertices of the trajectory with $j > i$.
Once that happens, we add the segment $p_ip_j$ to our simplification and set the vertex $p_j$ as the
next starting point.
We repeat the process until we reach the end of the trajectory.
This method can be used to obtain a simplification of complexity $\ell$ by performing a binary
search on the threshold value.

The Imai--Iri approach allows to compute an optimal simplification in a local sense, i.e.,
assuming the matching is fixed at the points that coincide between the trajectory and its
simplification.
There are two reasonable approaches to its implementation:
\begin{enumerate}
    \item Given a threshold, save all the possible line segments between vertices of the trajectory
    for which the distance to the corresponding subtrajectory is below the threshold.
    Then find the shortest path from the beginning to the end of the trajectory.
    This can be adapted to targeting the specific $\ell$ using binary search on the threshold value.
    \item Given a target $\ell$, use a dynamic program to record the costs of all possible shortcuts
    in $\ell$ steps and find the lowest-cost solution.
    This approach seems more suitable for sum-based distances like DTW and CDTW.
\end{enumerate}
It is not obvious which approach would be preferable in our setting, since any of these could be
used with DTW, the Fr\'echet distance, and CDTW.
We conduct an experimental evaluation and describe the results in Section~\ref{sec:simpl_exp}.
Based on these results, we choose the greedy approach with Fr\'echet distance in all our clustering
pipelines.

\subsection{Improving Cluster Centers}
Petitjean et al.~\cite{petitjean} introduced a method called \emph{DTW Baricenter Averaging (DBA)}
to improve the centers in $k$-means clustering using the DTW distance.
Let $\mathcal{C}$ be a trajectory cluster and let $C = (c_1, \dots, c_l)$ be its initial center.
The DBA method first computes the DTW matching between the center $C$ and all trajectories in the
cluster $\mathcal{C}$.
For all $1 \le i \le l$, let $\Tilde{c_i}$ be the average of all points matched to $c_i$.
We replace the initial center $C$ with the new center $\Tilde{C} = (\Tilde{c_1}, \dots, \Tilde{c_l})$
if it induces a lower cost.
The procedure is repeated until the new center does not improve upon the current center.

\begin{figure}
    \centering
    \includegraphics[width=\columnwidth]{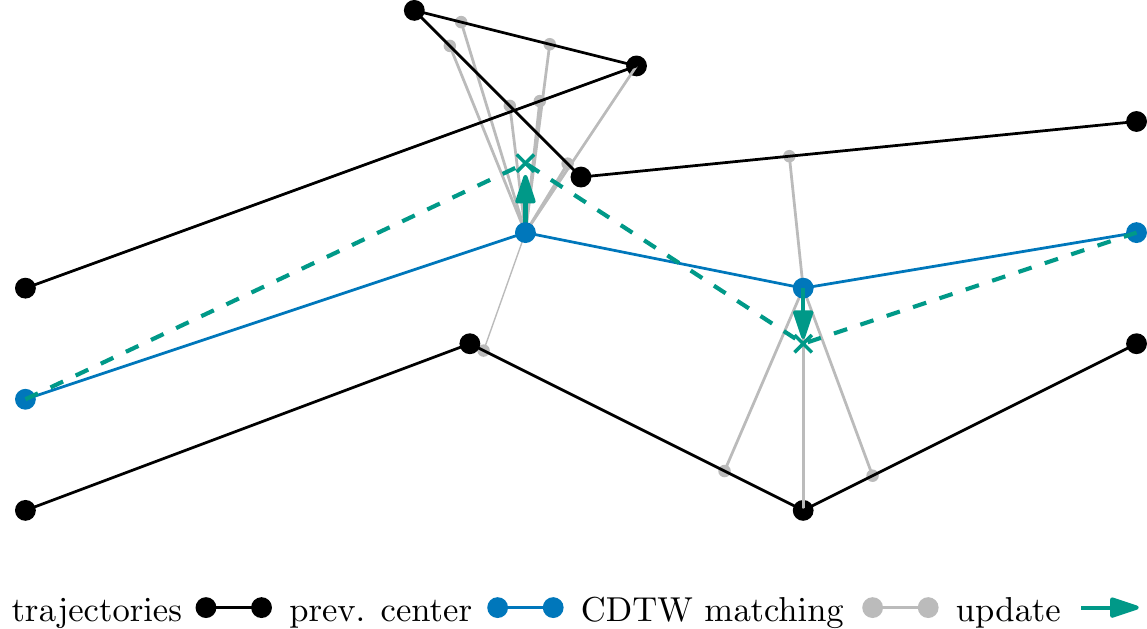}
    \Description{Two trajectories and the center.
    The vertices of the center are matched to subtrajectories.
    We sample those subtrajectories, thus pulling the vertices of the center more strongly in the
    direction where there are more samples.}
    \caption{The CDBA update method, with the CDTW matching sampled uniformly to compute the updated
    centers.}
    \label{fig:cdba_update}
\end{figure}

We use a similar method to improve the cluster centers arising from the CDTW distance.
We call our method \emph{CDTW Barycenter Averaging (CDBA).}
A key difference between the DBA method and our CDBA method is that ours uses a continuous matching,
so a vertex on $C$ may be mapped to an entire subtrajectory rather than a set of vertices.
In this situation, we uniformly sample points on the matched subtrajectory.
The number of sampled points is proportional to the ratio of the length of the subtrajectory to the
length of the trajectory.
See Figure~\ref{fig:cdba_update} for an example where the matching is sampled to compute the updated
centers.

The sampling method for vertices mapped to a continuous subtrajectory has been previously used by
Buchin et al.~\cite{buchin2019-sigspatial}.
Their work focuses on $(k, \ell)$-center clustering using the Fr\'echet distance, and we refer to
their method as \emph{free space averaging (FSA).}
In the FSA method, rather than updating vertices of the center trajectory to be the mean of the
points they are matched to, vertices are assigned to the center of the minimum enclosing circle of
the points they are matched to in the Fr\'echet matching.

The intuition behind the center update methods described above is that by moving vertices of the
center trajectory closer to the points on cluster trajectories to which they are aligned with
respect to the distance measure under consideration, we expect the new center trajectory to be a
better fit for the cluster.
This intuition makes the implicit assumption that the alignment between the center trajectory and
trajectories in the cluster does not change too much after the update.
The update method for each distance function also depends on the type of clustering considered to be
most natural for that particular distance function.
The Fr\'echet distance minimizes the maximum distance between aligned points.
Similarly, $k$-center clustering minimizes the maximum distance between each trajectory and the
trajectories in the associated cluster, and so $k$-center clustering seems a natural choice for
clustering with the Fr\'echet distance.
The FSA method described was designed by Buchin et al.~\cite{buchin2019-sigspatial} for the purpose
of improving centers in $(k, l)$-center clustering.
Both DTW and CDTW are distance measures that minimize the sum or integral of distances between
aligned points.
Therefore, $k$-means or $k$-medians clustering seem to be more natural choices for these distance
measures.
The DBA center update method we have described was designed by Petitjean et al.~\cite{petitjean} in
the context of $k$-means clustering under the DTW distance measure, while we apply the CDBA method
for $k$-medians clustering under CDTW.

The CDBA method is based on the strategy of moving vertices.
However, the definition of CDTW also takes into consideration the internal points on edges.
Therefore, we also consider a center update method which aims to optimize the position of edges,
rather than just vertices.
With this in mind, we introduce an alternative center update method for $(k, \ell)$-medians
clustering under CDBA which we call the \emph{wedge method.}
Given an initial center trajectory $C=(c_1, c_2, \dots, c_{\ell})$ and the associated cluster
$\mathcal{C}$, the wedge method works as follows.
For each $c_i$, $1 < i < \ell$, we define the line segments $s_L = c_{i - 1}c_i$ and
$s_R = c_ic_{i + 1}$.
We then define the sets $L$ and $R$ to be all vertices of trajectories in $\mathcal{C}$ aligned by
CDTW to $s_L$ and $s_R$, respectively.
For each vertex $p$ of a trajectory $P \in \mathcal{C}$ aligned to either $s_L$ or $s_R$ under CDTW,
we define $w(p)$ to be the sum of lengths of the segments preceding and following $p$ on $P$.
We search for a point $\tilde{c_i}$ by perturbing $c_i$ so as to minimize
\[\sum_{p \in L} w(p) \cdot \gamma(p, c_{i - 1}\tilde{c_i})^2 +
\sum_{p \in R} w(p) \cdot \gamma(p, \tilde{c_i}c_{i + 1})^2\,,\]
where $\gamma(p, s)$ denotes the distance from a point $p$ to the closest point on a line segment
$s$.

Finally, we can perform a similar optimization for the first and last points on the initial center
trajectory.
Depending on the nature of the data we are working on, it may make more sense to keep the start and
end points fixed, as is the case in the data set of pigeon flight paths we study.
We include the weight terms $w(\cdot)$ as we expect vertices between longer segments to contribute
more to the value of the CDTW distance.

\section{Experiments}\label{sec:experiments}
In this section, we describe the experiments performed on the methods described above and their
results.
\subsection{Data Sets}
We have conducted experiments on three different open-source real-world data sets.
The first of these is a set of handwritten characters of the English
alphabet~\cite{character-dataset} from the UCI Machine Learning Repository~\cite{dua:2019}.
The data set contains a few hundred examples in each of twenty classes, with each class
corresponding to a character that can written in a single stroke.

We also make use of the pigeon flight path trajectory data set by Mann et al.~\cite{mann2010}.
The data set consists of flight paths of pigeons released from four distinct release sites as they
fly to a common return site.
For each release site, the data set contains trajectories corresponding to seven or eight distinct
pigeons, each of which is released approximately twenty times.

Finally, we also make use of the stork migration data by Rotics et al.~\cite{movebank-paper}, made
available at the Movebank data repository~\cite{movebank-repository}.
The data set contains the movements of 35~adult storks over four years as they migrate from Africa
to Europe.

As our methods become impractical for very large trajectories (some trajectories in the stork
migration data set contain over 15000~points), we pre-process the data by regularly sampling points
to ensure the trajectories we use have at most a few hundred vertices.
The coordinates in the pigeon and stork data sets are given as latitude and longitude.
Since our methods assume trajectories in the plane, we apply projections to the data sets.
We use the transverse Mercator projection for the pigeon data set and the two-point equidistant
projection for the stork data set.

We conduct experiments to evaluate the three steps in our trajectory clustering implementation,
that is, our initial clustering, our trajectory simplification, and our center improvement methods.
We compare our results to existing approaches.
In particular, we compare the quality of trajectory simplifications based on DTW and the Fr\'echet
distance with our algorithm based on CDTW, and we compare the quality of center improvement methods
based on DBA and FSA with our CDBA and wedge methods.
We make the source code of our C++ implementation publicly available.\footnote{See
\url{https://github.com/Mesoptier/trajectory-clustering}.}

\subsection{Initial Clustering}
We apply the PAM clustering algorithm to the pigeon flight data.
For each of the four release sites, we compute $(k, \ell)$-clusterings, for $k \in \{1, \dots, 7\}$
and $\ell = 10$, using the following algorithms with CDTW and greedy simplification with the
Fr\'echet distance:
\begin{enumerate}
    \item The Gonzalez algorithm;
    \item PAM after initializing the centers using the Gonzalez algorithm;
    \item PAM after greedily initializing the centers according to the $\phi_1$ cost function.
\end{enumerate}

Since the Gonzalez algorithm is randomized, when clustering via algorithms~(1) and~(2), we take the
best of five iterations as the final result.
To compare the approaches, we evaluate the values of the $(k, \ell)$-medians scores for the
resulting clusterings.
Both PAM and PAM with Gonzalez initialization consistently outperform the vanilla Gonzalez approach.
This is to be expected, since the Gonzalez algorithm does not aim to optimize the
$(k, \ell)$-medians score.
The results also indicate that PAM and PAM with Gonzalez initialization often arrive at the same
final clustering.

\subsection{Trajectory Simplification}\label{sec:simpl_exp}
As described in Section~\ref{sec:simpl_desc}, there are many possible approaches to trajectory
simplification in our setting.
We focus on the greedy approach and the dynamic programming approach, using DTW, the Fr\'echet
distance, and CDTW.
We experimentally evaluate these approaches to gauge their performance both visually and with
respect to CDTW.
In order to do that, we sample trajectories from two of the data sets we use, the character and the
pigeon data set.
We additionally subsample each trajectory at regular intervals to get trajectories of complexity~50.
We then use the different methods to simplify the trajectories to complexity~12 and evaluate the
resulting simplifications with respect to CDTW to the original trajectories.
The results are shown in Tables~\ref{tab:simpl_char} and~\ref{tab:simpl_pigeon}.

\begin{table}
\caption{Simplification methods on the characters data set.
Reported values are computed over the data set w.r.t.\@ CDTW between the trajectories and their
simplifications.}
\centering
\begin{tabular}{l l r r r}
\toprule
          &           &  mean &  min &    max\\
\midrule
DTW       & Greedy    & 255.5 & 1.90 & 2001.0\\
          & Imai--Iri & 175.6 & 1.58 & 1827.4\\
Fr\'echet & Greedy    &  71.4 & 0.88 &  695.2\\
          & Imai--Iri &  67.4 & 0.69 &  689.5\\
CDTW      & Greedy    &  68.2 & 0.93 &  665.5\\
          & Imai--Iri &  36.9 & 0.47 &  469.1\\
\bottomrule
\end{tabular}
\label{tab:simpl_char}
\end{table}

\begin{table}
\caption{Simplification methods on the pigeons data set (the Bladon Heath site).
Reported values are computed over the data set w.r.t.\@ CDTW between the trajectories and their
simplifications.}
\centering
\begin{tabular}{l l r r r}
\toprule
          &           &  mean &  min &    max\\
\midrule
DTW       & Greedy    & 0.829 & 0.003 &  9.87\\
          & Imai--Iri & 0.586 & 0.005 &  4.68\\
Fr\'echet & Greedy    & 0.879 & 0.024 & 10.19\\
          & Imai--Iri & 0.676 & 0.023 &  6.60\\
CDTW      & Greedy    & 0.883 & 0.008 &  9.14\\
          & Imai--Iri & 0.335 & 0.005 &  3.87\\
\bottomrule
\end{tabular}
\label{tab:simpl_pigeon}
\end{table}

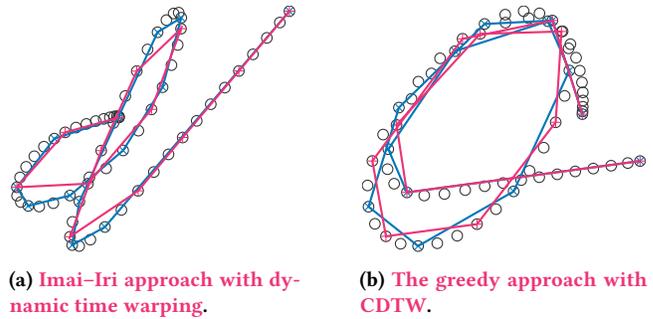
\begin{figure}
\centering
\begin{minipage}[b]{.45\columnwidth}
\centering
\tikzsetnextfilename{ii_dtw_ii_cdtw}
\begin{tikzpicture}
\begin{axis}[set layers,mark layer=axis background]
\addplot[black!80,only marks,mark=o] table {data/simpl/crv52};
\addplot[c2,mark=x] table {data/simpl/best52};
\addplot[c4,mark=+] table {data/simpl/ii_dtw52};
\end{axis}
\end{tikzpicture}
\subcaption{\textcolor{c4}{Imai--Iri approach with dynamic time warping}.}
\label{fig:simpl_dtw}
\end{minipage}\hfill
\begin{minipage}[b]{.45\columnwidth}
\centering
\tikzsetnextfilename{gr_cdtw_ii_cdtw}
\begin{tikzpicture}
\begin{axis}[set layers,mark layer=axis background]
\addplot[black!80,only marks,mark=o] table {data/simpl/crv154};
\addplot[c2,mark=x] table {data/simpl/best154};
\addplot[c4,mark=+] table {data/simpl/gr_cdtw154};
\end{axis}
\end{tikzpicture}
\subcaption{\textcolor{c4}{The greedy approach with CDTW}.}
\label{fig:simpl_cdtw}
\end{minipage}
\Description{Letters `l' and `o', handwritten, and the simplifications.
The Imai--Iri approach with DTW misses curved parts.
The greedy approach with CDTW is good, although uses the placement of points less nicely than
Imai--Iri with CDTW.}
\caption{Example simplifications for DTW and CDTW.
Over the original trajectory, two simplifications are shown in color.
\textcolor{c2}{Imai--Iri approach with CDTW} performs best.}
\label{fig:simpl1}
\end{figure}

As can be seen from the tables, using CDTW with Imai--Iri gives the best results, as expected.
Unfortunately, this is also the slowest method, several orders of magnitude slower in our
experiments than approaches that use the Fr\'echet distance or DTW, which makes it impractical.
Using the greedy approach with CDTW gives simplifications of only slightly worse quality (see
Figure~\ref{fig:simpl_cdtw}), but still shows high computation times.

Other contenders show comparable (amongst each other) results on the two data sets, as can be seen
from the results in both tables.
They are also all much faster; so, it makes sense to examine the specific defects that arise from
each approach to better understand which methods are more suitable in our setting.
On the characters data set, DTW fares poorly both using Imai--Iri and the greedy approach.
Figure~\ref{fig:simpl_dtw} showcases a bad example for dynamic time warping with the Imai--Iri
approach, compared to the performance of Imai--Iri with CDTW on the same trajectory.
Due to the discrete nature of DTW, points have to be placed somewhat regularly, including on
straight-line parts of the trajectory.
Since $\ell$ is fixed, DTW-based approaches visually perform poorly on the curved parts.
This is reflected in the values of CDTW.
Even though DTW performs better on (part of) the pigeon data set, using it in general settings is
not ideal, since there are many real-world trajectories with both long straight segments and smooth
curvature that will yield bad results, e.g., plane or road vehicle trajectories.

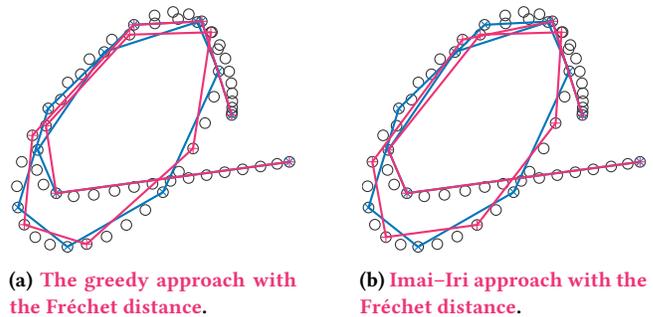
\begin{figure}
\centering
\begin{minipage}[b]{.45\columnwidth}
\centering
\tikzsetnextfilename{gr_fr_ii_cdtw}
\begin{tikzpicture}
\begin{axis}[set layers,mark layer=axis background]
\addplot[black!80,only marks,mark=o] table {data/simpl/crv154};
\addplot[c2,mark=x] table {data/simpl/best154};
\addplot[c4,mark=+] table {data/simpl/gr_fr154};
\end{axis}
\end{tikzpicture}
\subcaption{\textcolor{c4}{The greedy approach with the Fr\'echet distance}.}
\label{fig:simpl_grfr}
\end{minipage}\hfill
\begin{minipage}[b]{.45\columnwidth}
\centering
\tikzsetnextfilename{ii_fr_ii_cdtw}
\begin{tikzpicture}
\begin{axis}[set layers,mark layer=axis background]
\addplot[black!80,only marks,mark=o] table {data/simpl/crv154};
\addplot[c2,mark=x] table {data/simpl/best154};
\addplot[c4,mark=+] table {data/simpl/ii_fr154};
\end{axis}
\end{tikzpicture}
\subcaption{\textcolor{c4}{Imai--Iri approach with the Fr\'echet distance}.}
\label{fig:simpl_iifr}
\end{minipage}
\Description{Letter `o', handwritten, and the simplifications.
Both approaches with the Fr\'echet distance deviate somewhat on the long segments, but perform
surprisingly well.}
\caption{Example simplifications for the Fr\'echet distance.
Over the original trajectory, two simplifications are shown in color.
\textcolor{c2}{Imai--Iri approach with CDTW} performs best.}
\label{fig:simpl2}
\end{figure}

Approaches based on the Fr\'echet distance fare much better on the character data set and comparably
on the pigeons data set.
Visual representations of the worst examples compared to Imai--Iri CDTW are shown in
Figure~\ref{fig:simpl2}.
Note that here the issue is simply the lack of attention for long segments that do not quite
coincide with the trajectory, something that is cost-free for the Fr\'echet distance, but penalized
quite heavily for CDTW.
Fr\'echet-induced simplifications, however, do concentrate the points in the curved sections of the
input instead of the straight-line segments, and tend to use less points than the available number
$\ell$.
Note that we use the simplifications of the center trajectories only as a starting point in our
clustering approach before doing further center improvements.
Our center improvement strategies can add points in the places that benefit most from it, thus
alleviating some of the issues with the Fr\'echet simplifications.
So, Fr\'echet distance-based simplifications seem to be a good fit for our clustering approach,
assuming we cannot use CDTW-based simplification due to the high running time.
The greedy approach using the Fr\'echet distance is very fast, on par with DTW computations, so
using it on the fly is feasible.
Therefore, in our clustering approach we use the greedy approach with the Fr\'echet distance to
compute the $\ell$-simplifications.

\subsection{Improving Cluster Centers}
We conduct experiments to evaluate the quality of the DBA, FSA, CDBA, and wedge methods for
improving cluster centers.
First, we conduct an experiment using the character data set to compare the quality of the centers
produced according to the CDTW $(k, \ell)$-medians cost.
In particular, since our CDBA and wedge methods are designed to minimize the CDTW
$(k, \ell)$-medians cost, we expect them to outperform DBA and FSA.
The experiment is conducted as follows.
For each character, we sample fifty trajectories from the data.
For each of these sets of fifty trajectories, we compute an initial $(k, \ell)$-medians clustering,
for a fixed $k$ and a range of values for $\ell$, using our version of the PAM method with CDTW and
greedy simplification with the Fr\'echet distance.
We then apply four different center improvement algorithms to the same initial clustering.
Each method iteratively updates the center trajectory until the method does not make an improvement
or until we reach twenty iterations.

We manually select the values of $k$ and $\ell$ that reasonably reflect the underlying trajectories
in the data set.
For the characters data set, there seems no inherently correct value for $k$, so we choose $k = 2$
to accommodate for possible deviations in handwriting.
Furthermore, we choose $\ell \in \{6, 9, 12\}$, as the shape of the characters can be represented by
trajectories of this complexity.
Note that the general problem of parameter selection is a difficult one.
Reddy and Vinzamuri~\cite{aggarwal2013} discuss common ways to choose $k$.
The general model selection techniques suitable for picking $k$ (e.g., information criteria) can
also be applied to choose $\ell$.
These methods are especially preferred for data sets where the true complexities of the underlying
trajectories are difficult to estimate.

\begin{figure}
    \centering
    \includegraphics[width=\linewidth]{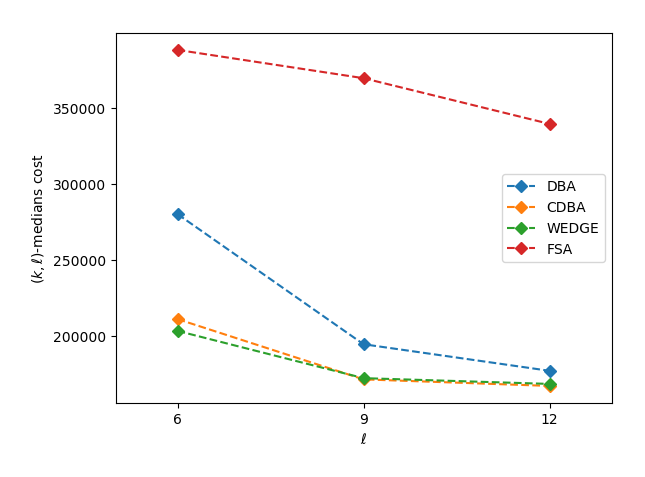}
    \Description{A line plot with $k = 6, 9, 12$ on the $x$-axis and the cost on the $y$-axis.
    The FSA line is high above the rest.
    The wedge method is only slightly better than CDBA.
    DBA is comparable to both for $k = 12$, but becomes much worse for $k = 6$.}
    \caption{A comparison of the average $(k,\ell)$-medians score over the twenty characters, for
    different values of $\ell$.}
    \label{fig:vary_ell}
\end{figure}

The results of the experiments for $k = 2$ and $\ell \in \{6, 9, 12\}$ are shown in
Figure~\ref{fig:vary_ell}.
On the $y$-axis are the average CDTW $(k, \ell)$-medians scores over the twenty characters in the
data sets.
The FSA method obtains significantly higher $(k, \ell)$-medians scores for all values of $\ell$.
The DBA method obtains lower $(k, \ell)$-medians scores than FSA, but the score degrades more
quickly when the complexity of the cluster center decreases.
This supports our claim in the introduction that clustering based on DTW is sensitive to
low-complexity cluster centers.

\begin{figure}
    \centering
    \includegraphics[width=\linewidth]{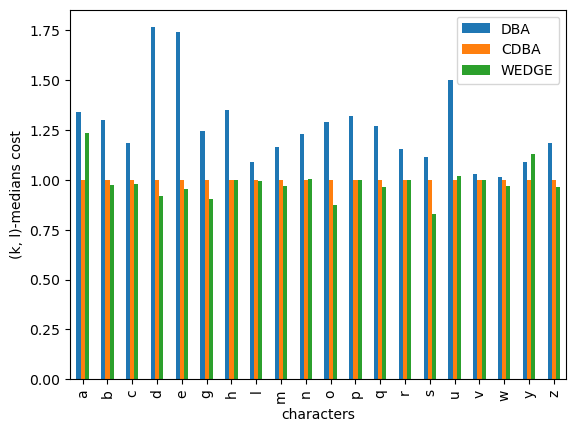}
    \Description{A column plot per character.
    CDBA cost is scaled to $1$; the wedge cost is mostly a bit lower, the DBA cost is mostly a bit
    higher, with quite a few columns where it is significantly higher.}
    \caption{Comparison of improvement methods on handwritten character data set with $k = 2$,
    $\ell = 6$.}
    \label{fig:char_exp_6_4}
\end{figure}

Next, we plot the CDTW $(k, \ell)$-medians cost of the twenty character classes in
Figure~\ref{fig:char_exp_6_4}.
We do so for $k = 2$ and $\ell = 6$.
The costs for FSA are significantly higher than the other three methods so we omit the FSA results
from the plots.
As expected, the CDBA and wedge methods outperform DBA on the CDTW $(k, \ell)$-medians cost on all
characters.
We select a few examples to highlight why we believe this is the case.
In particular, we show that there are apparent visual artifacts in clusterings where DBA and FSA
obtain a significantly higher $(k, \ell)$-medians cost.

In Figure~\ref{fig:k2_h} in the appendix, FSA obtains a significantly higher $(k, \ell)$-medians
cost than DBA, CDBA and the wedge method.
The green center computed by FSA is significantly distorted while the CDBA and DBA centers are more
consistent with the data.
In Figure~\ref{fig:k2_e} in the appendix, DBA obtains a significantly higher $(k, \ell)$-medians
cost than CDBA and the wedge method.
The purple center trajectory computed by DBA seems ``collapsed'' compared to the centers computed by
CDBA and the wedge method.
These visual artifacts are similar to those shown in Figures~\ref{fig:dtw_clustering_artifact}
and~\ref{fig:frechet_clustering_artifact}.

We perform similar experiments for the pigeons data set and the storks migration data set.
For both data sets we selected $k = 3$.
For the pigeon data we used $\ell = 10$, while for the stork data we used $\ell = 14$.
The $(k, \ell)$-medians costs for FSA and DBA were significantly higher than those for CDBA and the
wedge method, but figures for these were omitted due to space constraints.
We focus on highlighting the visual artifacts in the clustering that we believe cause FSA and DBA to
obtain these significant costs.
In Figure~\ref{fig:brc_plot} in the appendix, the artifacts of DBA are again clearly visible.
The blue center produced by FSA seems to zigzag more than the centers produced by CDBA and the wedge
method.
The CDTW-based center improvement methods appear to give the smoothest and most natural blue
center trajectories in these examples.
In Figure~\ref{fig:movebank} in the appendix, the blue centers produced by the CDBA and wedge
methods are smoother and a better fit for the data than the center produced by DBA.
The center trajectories produced by FSA bypass a sharp turn which appears to be a key feature of the
data set.
The CDBA and wedge methods nicely capture this feature.

Because of the visual artifacts in the characters, pigeons, and stork migration data sets, we are
inclined to believe that CDBA and the wedge method produce superior cluster centers.
For the same reason, we believe that the $(k, \ell)$-medians cost is a reasonable measure to
evaluate clustering algorithms, as it produces high costs on visually unintuitive cluster centers.

\section{Conclusions}
In this work we presented the first Continuous Dynamic Time Warping (CDTW) implementation that we
know of.
One reason why we are interested in a practical implementation of this distance measure is enabling
a center-based clustering approach that neither requires high-complexity representatives nor is
sensitive to outliers.
We conducted extensive experiments to evaluate our approach.

Our findings are that we should use the PAM method (partitioning around medoids) alongside the
Gonzalez algorithm for initializing the centers.
To compute an $\ell$-simplification of the initial centers, the Imai--Iri approach using CDTW gives
the best results, but the greedy approach using the Fr\'echet distance gives the best running time
performance.
We provided two novel methods of updating the initial cluster centers using the CDTW measure.
Our CDTW-based update method avoids many of the artifacts that appear in DTW- or Fr\'echet-based
update methods, especially in the presence of outliers or low-complexity centers.

Interesting future work includes finding algorithmic approaches that lead to faster running times
for CDTW computations and extending the usage of this distance measure to fields where DTW is
currently predominantly used.

\begin{acks}
The work of Aleksandr Popov is funded by the
\grantsponsor{NWO}{Dutch Research Council (NWO)}{https://www.nwo.nl/} under the project number
\grantnum{NWO}{612.001.801}.
\end{acks}
\raggedbottom\balance
\bibliographystyle{ACM-Reference-Format}
\bibliography{references}

\appendix\flushbottom\onecolumn
\section{Additional Figures}
\begin{figure}[b]
    \begin{minipage}[b]{0.3\textwidth}
        \includegraphics[width=\textwidth,trim={1.8cm 0.5cm 1.8cm 0.5cm},clip]{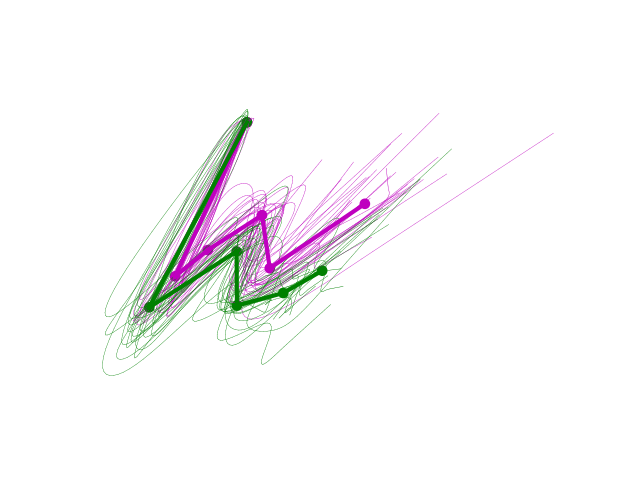}
        \subcaption{DBA.}
    \end{minipage}
    \hfill
    \begin{minipage}[b]{0.3\textwidth}
        \includegraphics[width=\textwidth,trim={1.8cm 0.5cm 1.8cm 0.5cm},clip]{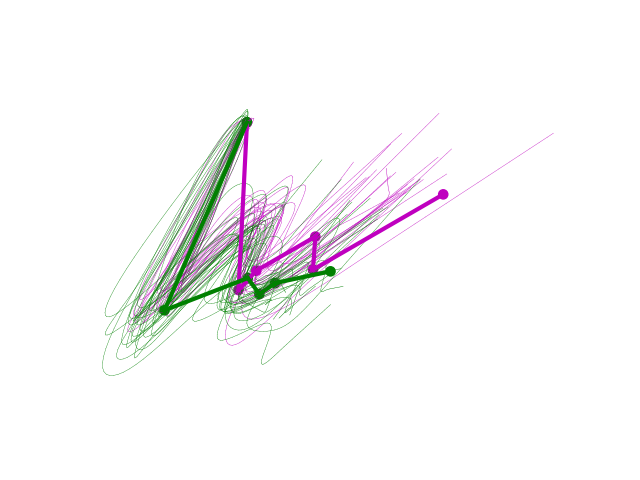}
        \subcaption{FSA.}
    \end{minipage}
    \hfill
    \begin{minipage}[b]{0.3\textwidth}
        \includegraphics[width=\textwidth,trim={1.8cm 0.5cm 1.8cm 0.5cm},clip]{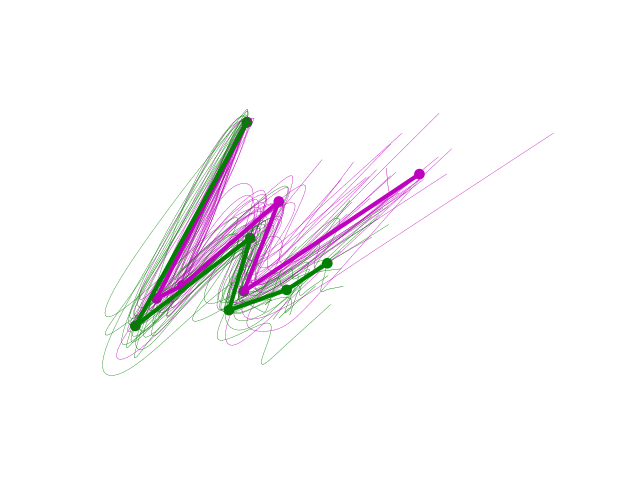}
        \subcaption{CDBA.}
    \end{minipage}
    \Description{Two clusters for character `h'.
    The DBA and CDBA cluster centers look like an `h', the FSA cluster centers have the second part
    of `h' barely pronounced.}
    \caption{(2, 6)-clusterings of \emph{h} characters computed with DBA, FSA and CDBA.}
    \label{fig:k2_h}
\end{figure}

\begin{figure}[b]
    \begin{minipage}[b]{0.3\textwidth}
        \includegraphics[width=\textwidth,trim={1.8cm 0.5cm 1.8cm 0.5cm},clip]{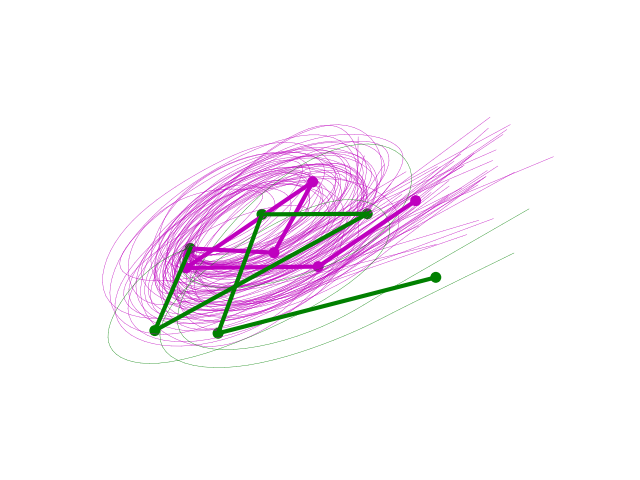}
        \subcaption{DBA.}
    \end{minipage}
    \hfill
    \begin{minipage}[b]{0.3\textwidth}
        \includegraphics[width=\textwidth,trim={1.8cm 0.5cm 1.8cm 0.5cm},clip]{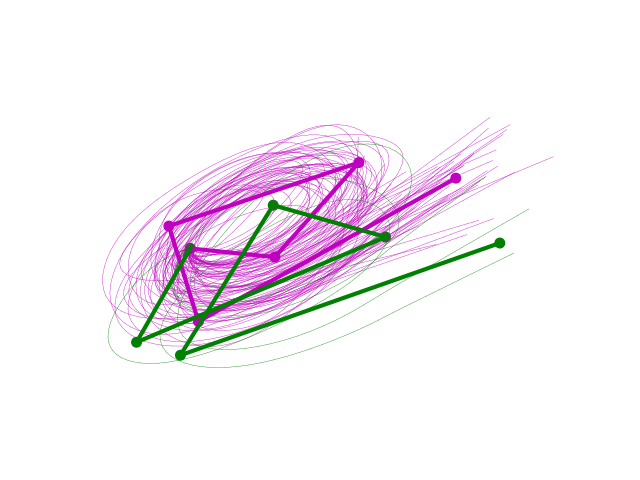}
        \subcaption{CDBA.}
    \end{minipage}
    \hfill
    \begin{minipage}[b]{0.3\textwidth}
        \includegraphics[width=\textwidth,trim={1.8cm 0.5cm 1.8cm 0.5cm},clip]{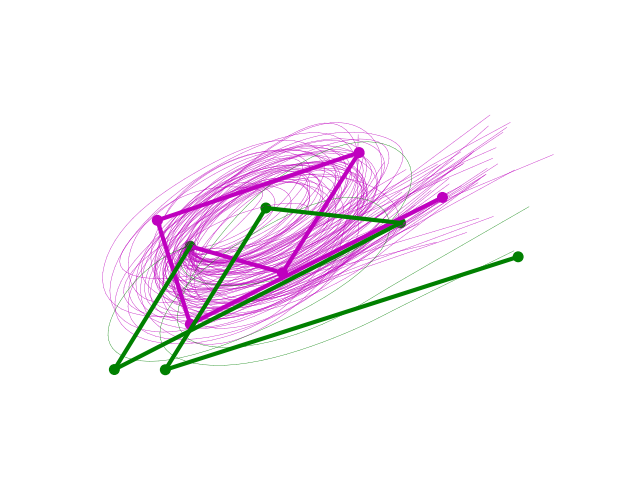}
        \subcaption{Wedge method.}
    \end{minipage}
    \Description{Clusters for difficult `e' characters.
    None of the centers are very clear, but the DBA centers are much more collapsed towards the
    center and less recognizable than the CDBA and wedge method centers.}
    \caption{(2, 6)-clusterings of \emph{e} characters computed with DBA, CDBA and the wedge
    method.}
    \label{fig:k2_e}
\end{figure}

\begin{figure}
    \begin{minipage}[b]{0.24\textwidth}
        \includegraphics[width=\textwidth,trim={3.8cm 1cm 3.8cm 1cm},clip]{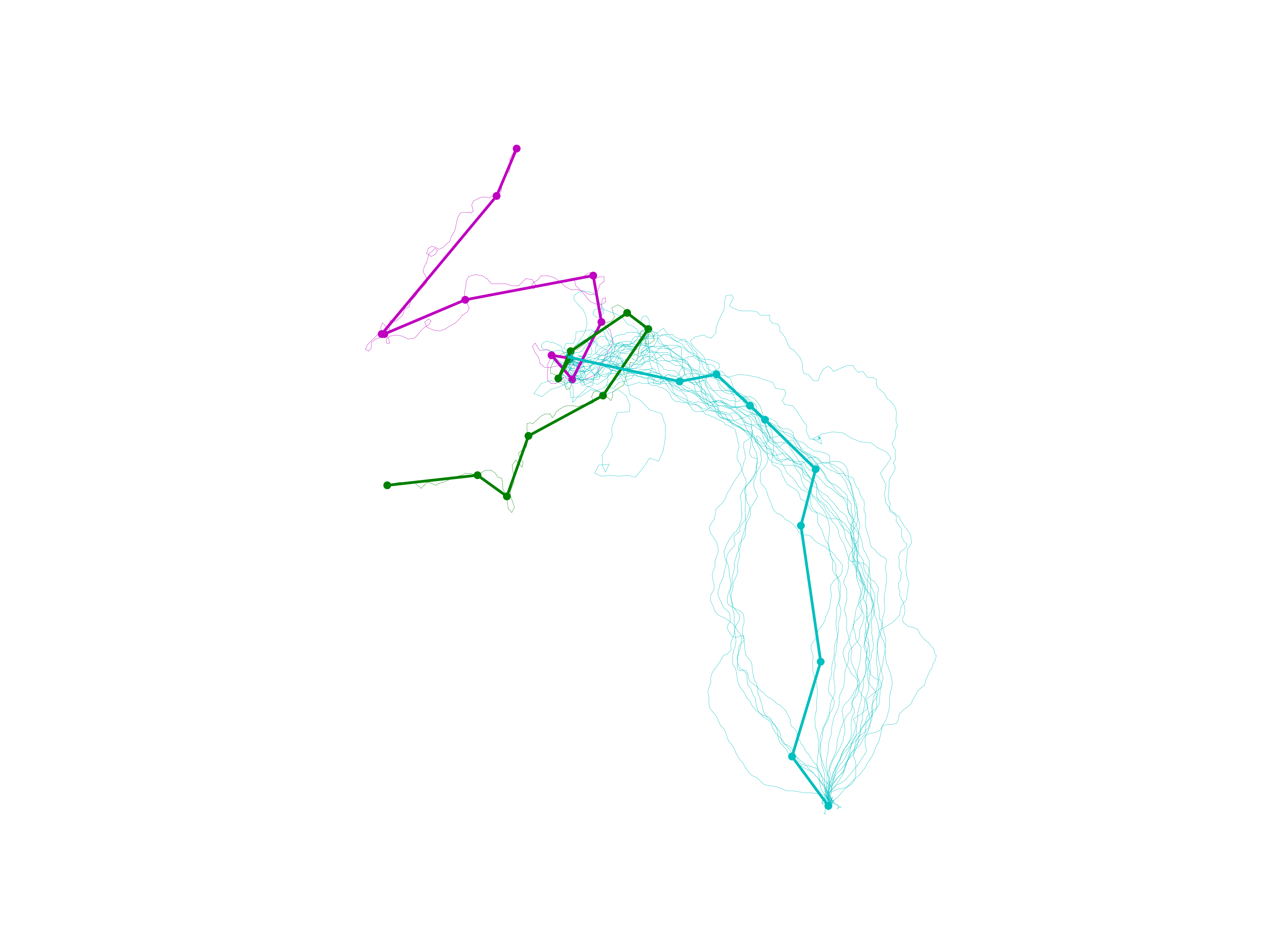}
        \subcaption{FSA.}
    \end{minipage}
    \hfill
    \begin{minipage}[b]{0.24\textwidth}
        \includegraphics[width=\textwidth,trim={3.8cm 1cm 3.8cm 1cm},clip]{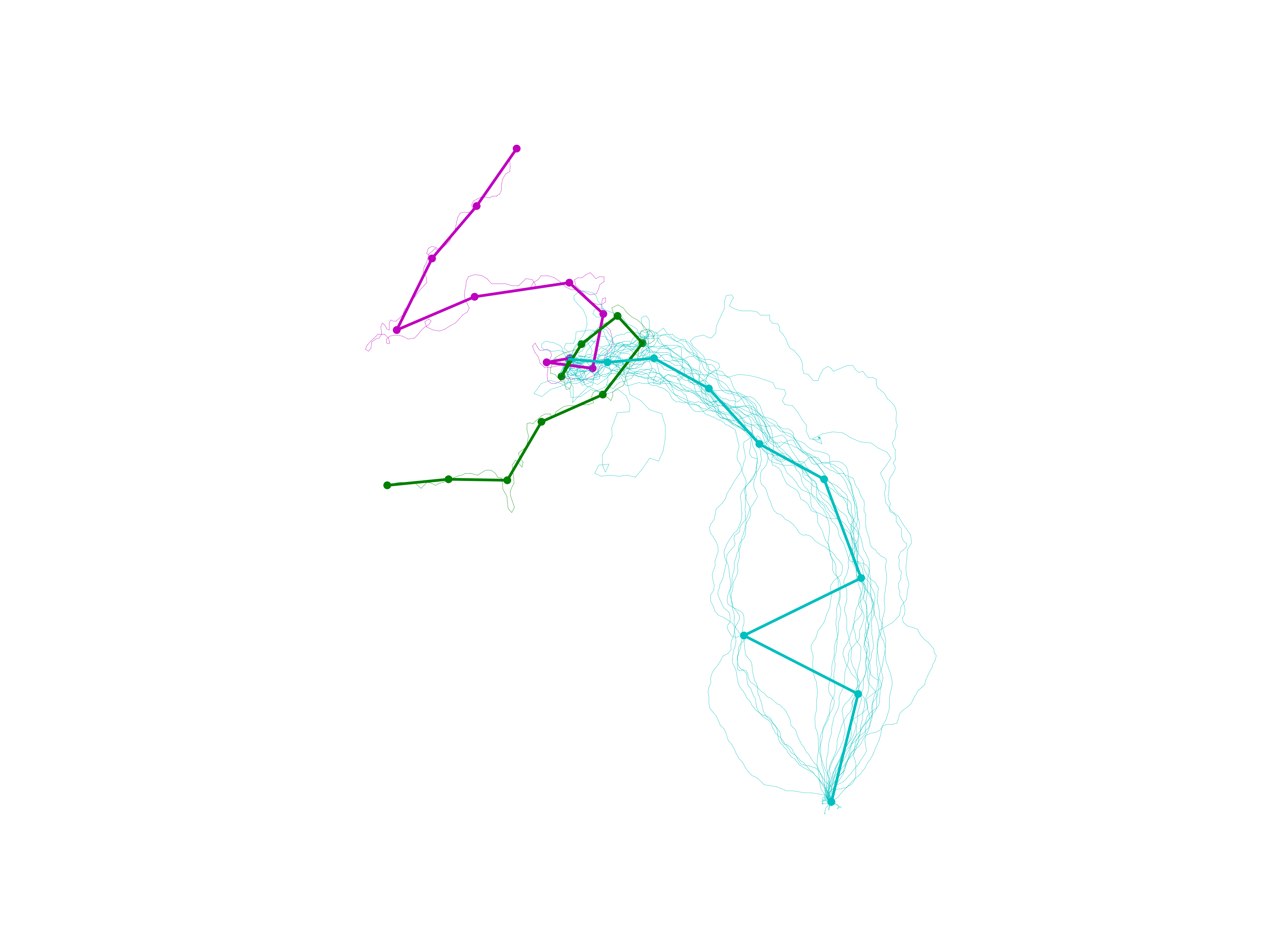}
        \subcaption{DBA.}
    \end{minipage}
    \hfill
    \begin{minipage}[b]{0.24\textwidth}
        \includegraphics[width=\textwidth,trim={3.8cm 1cm 3.8cm 1cm},clip]{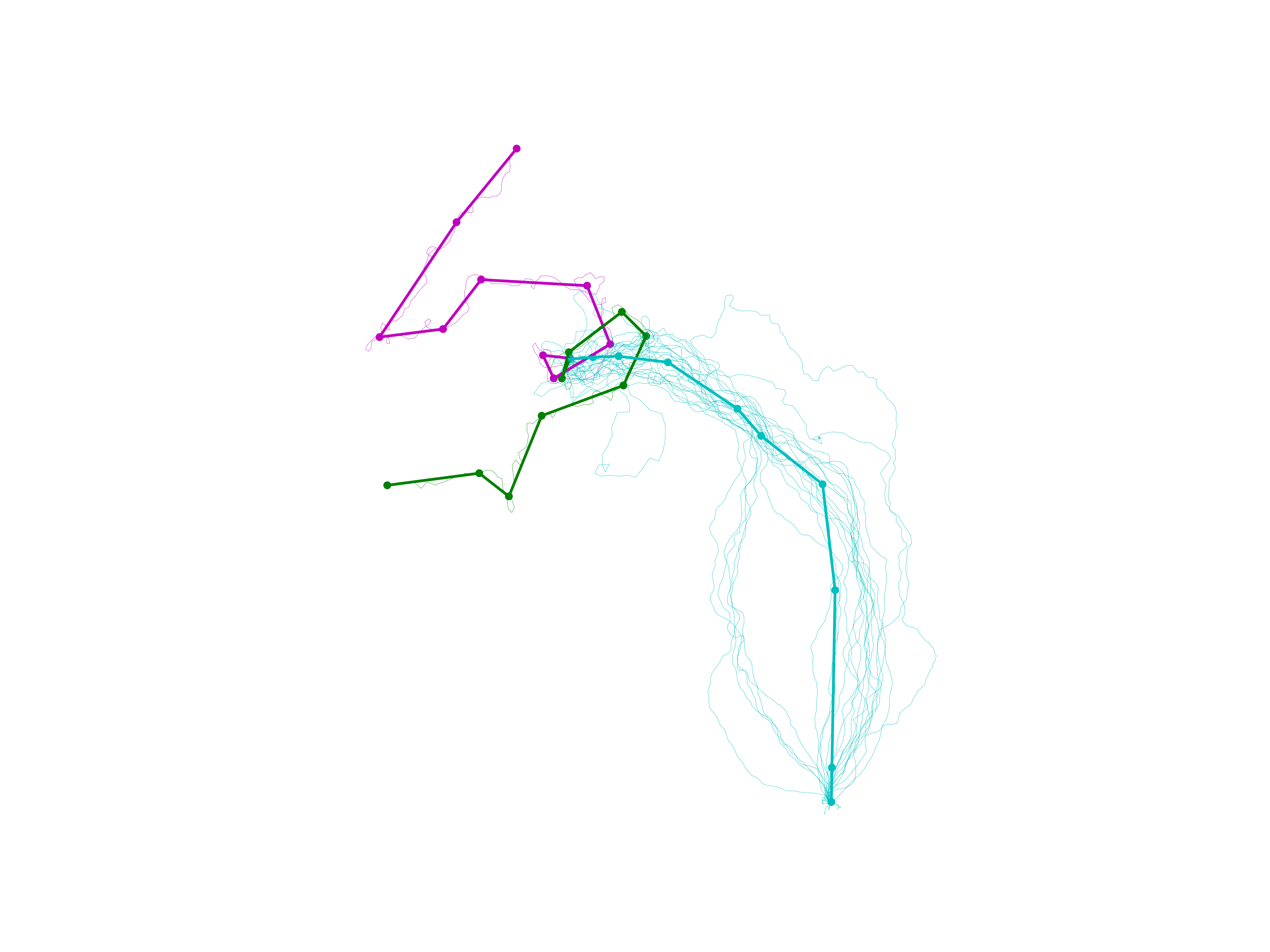}
        \subcaption{CDBA.}
    \end{minipage}
    \hfill
    \begin{minipage}[b]{0.24\textwidth}
        \includegraphics[width=\textwidth,trim={3.8cm 1cm 3.8cm 1cm},clip]{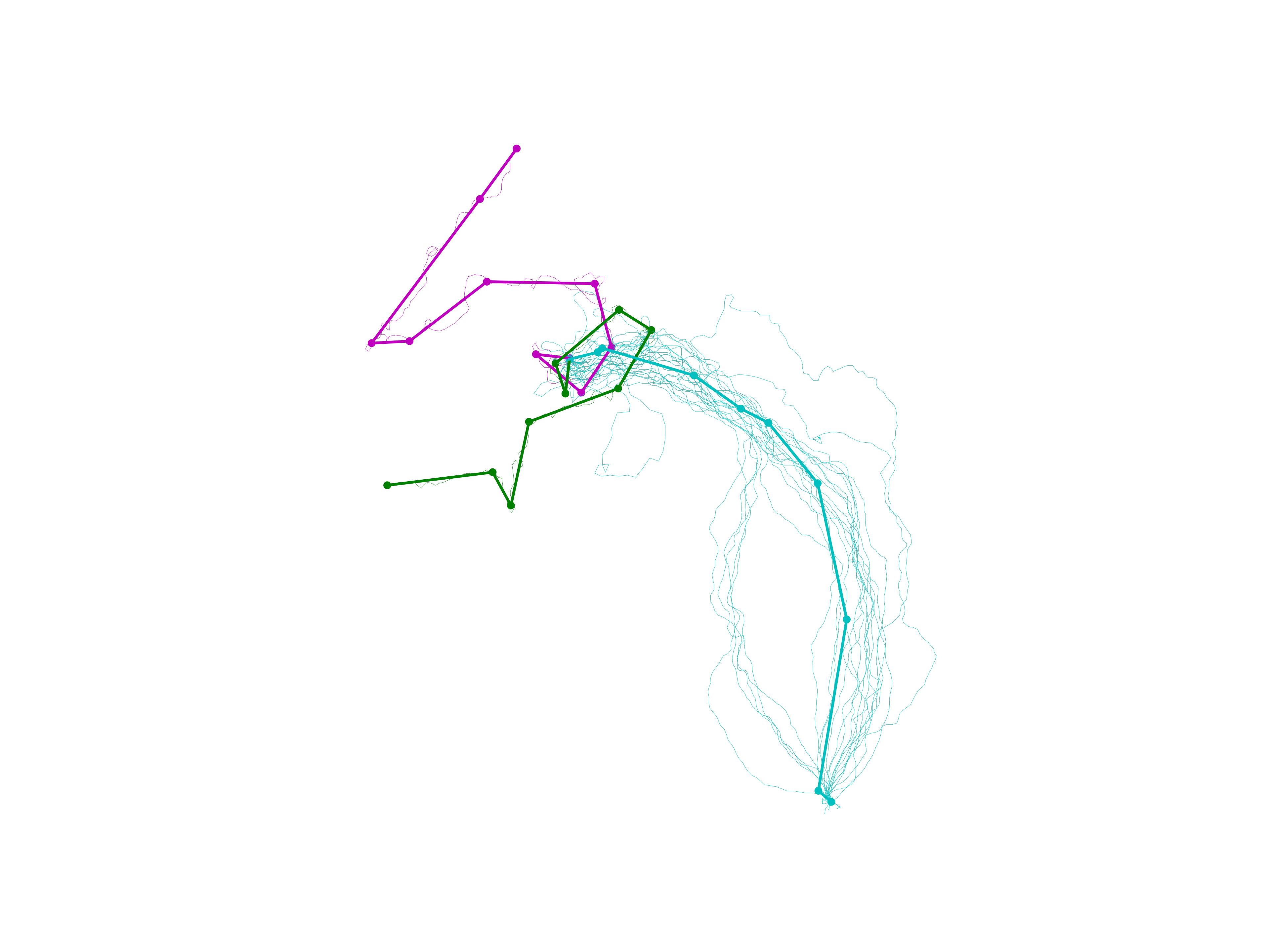}
        \subcaption{Wedge method.}
      \end{minipage}
      \Description{The three clusters on pigeon data. FSA has deviations, DBA has a clear jump to
      one of the trajectories.
      CDBA and the wedge method produce comparable results that look mostly smoother.}
      \caption{Centers of an initial clustering updated with (from left to right) FSA, DBA, CDBA,
      and the wedge method.
      The trajectory data used is the \emph{brc} pigeon from the pigeon data set.}
      \label{fig:brc_plot}
\end{figure}

\begin{figure}
    \begin{minipage}[b]{0.24\textwidth}
        \includegraphics[width=\textwidth,trim={3.8cm 1cm 3.8cm 1cm},clip]{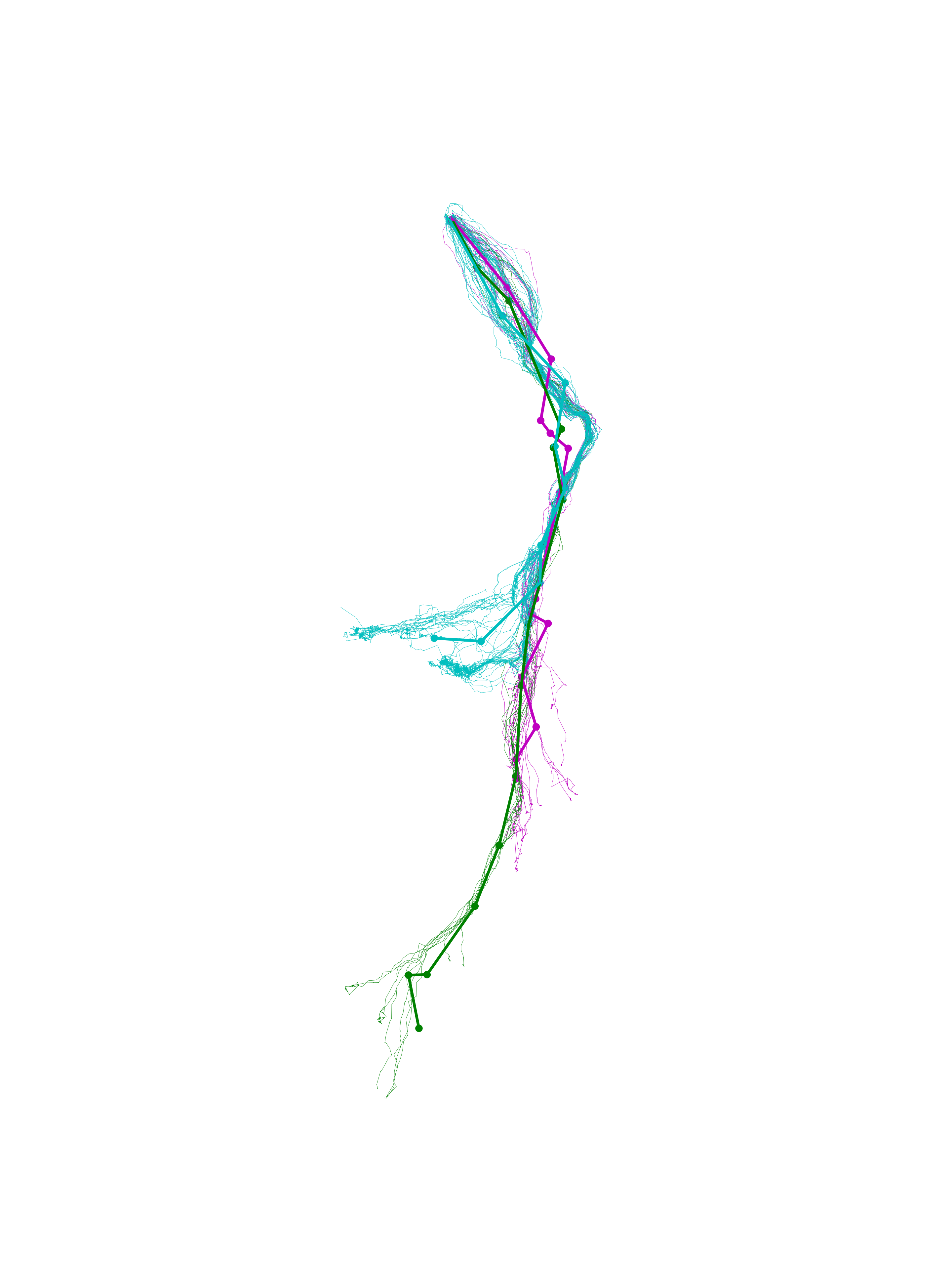}
        \subcaption{FSA.}
    \end{minipage}
    \hfill
    \begin{minipage}[b]{0.24\textwidth}
        \includegraphics[width=\textwidth,trim={3.8cm 1cm 3.8cm 1cm},clip]{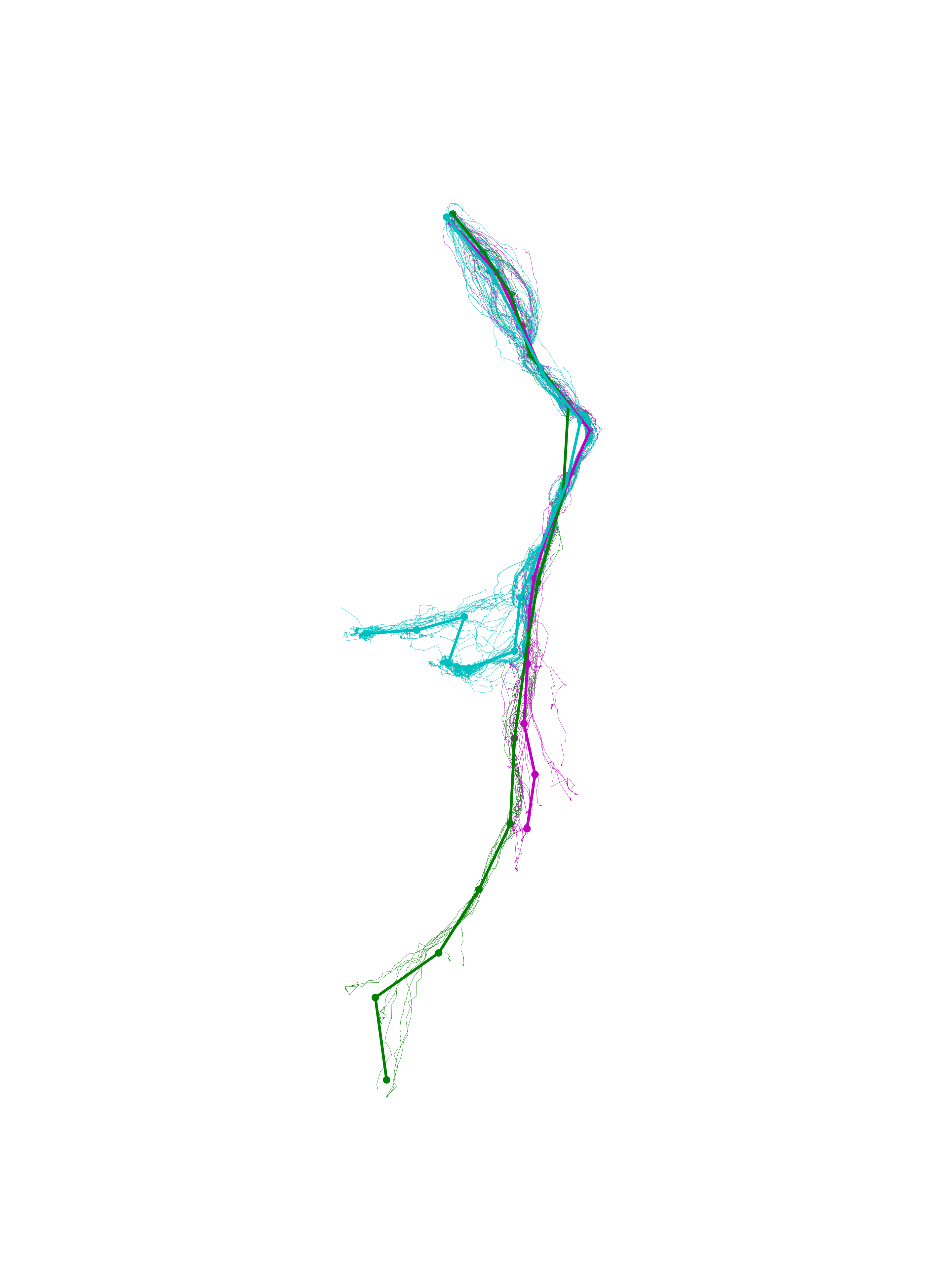}
        \subcaption{DBA.}
    \end{minipage}
    \hfill
    \begin{minipage}[b]{0.24\textwidth}
        \includegraphics[width=\textwidth,trim={3.8cm 1cm 3.8cm 1cm},clip]{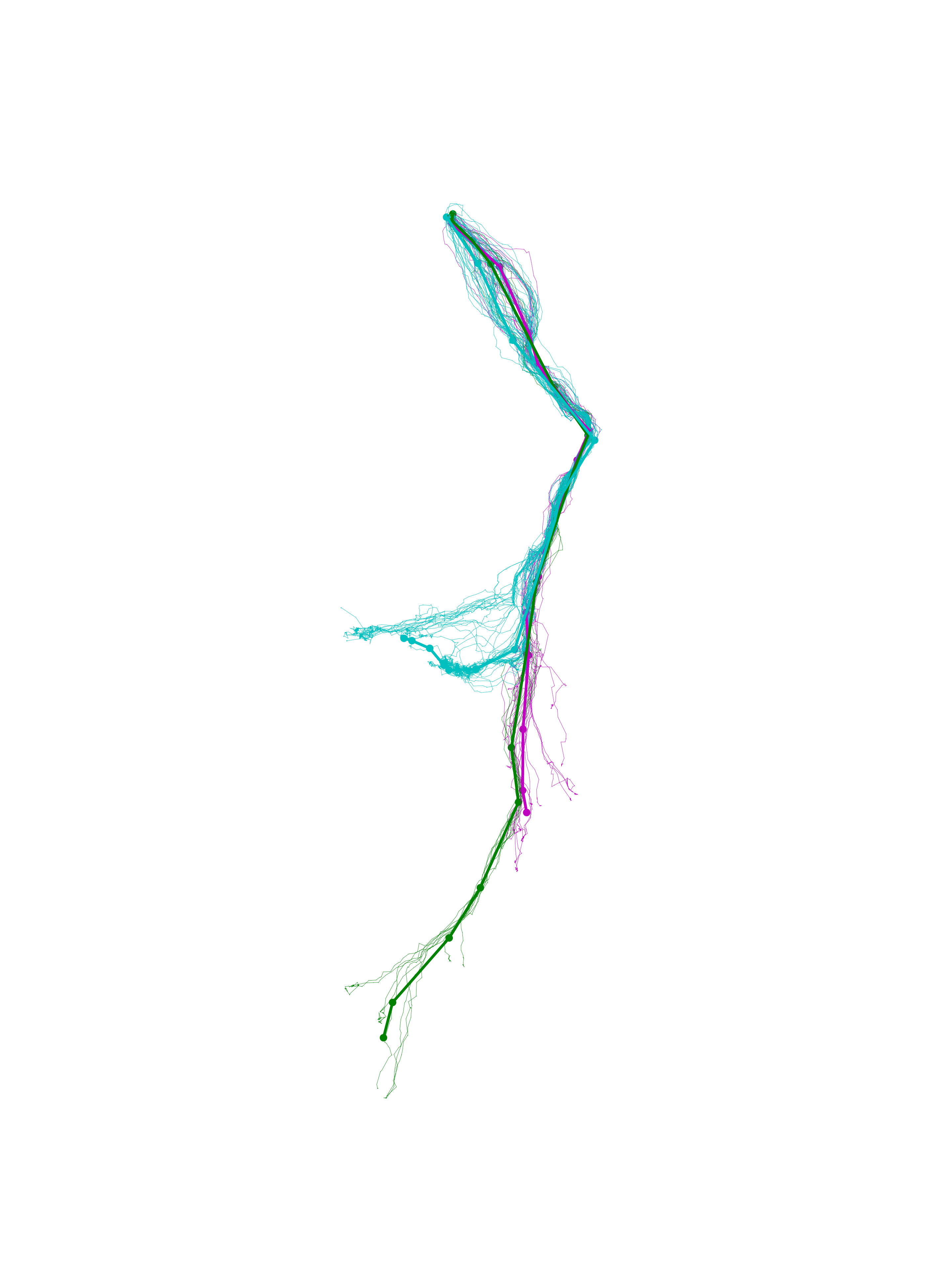}
        \subcaption{CDBA.}
    \end{minipage}
    \hfill
    \begin{minipage}[b]{0.24\textwidth}
        \includegraphics[width=\textwidth,trim={3.8cm 1cm 3.8cm 1cm},clip]{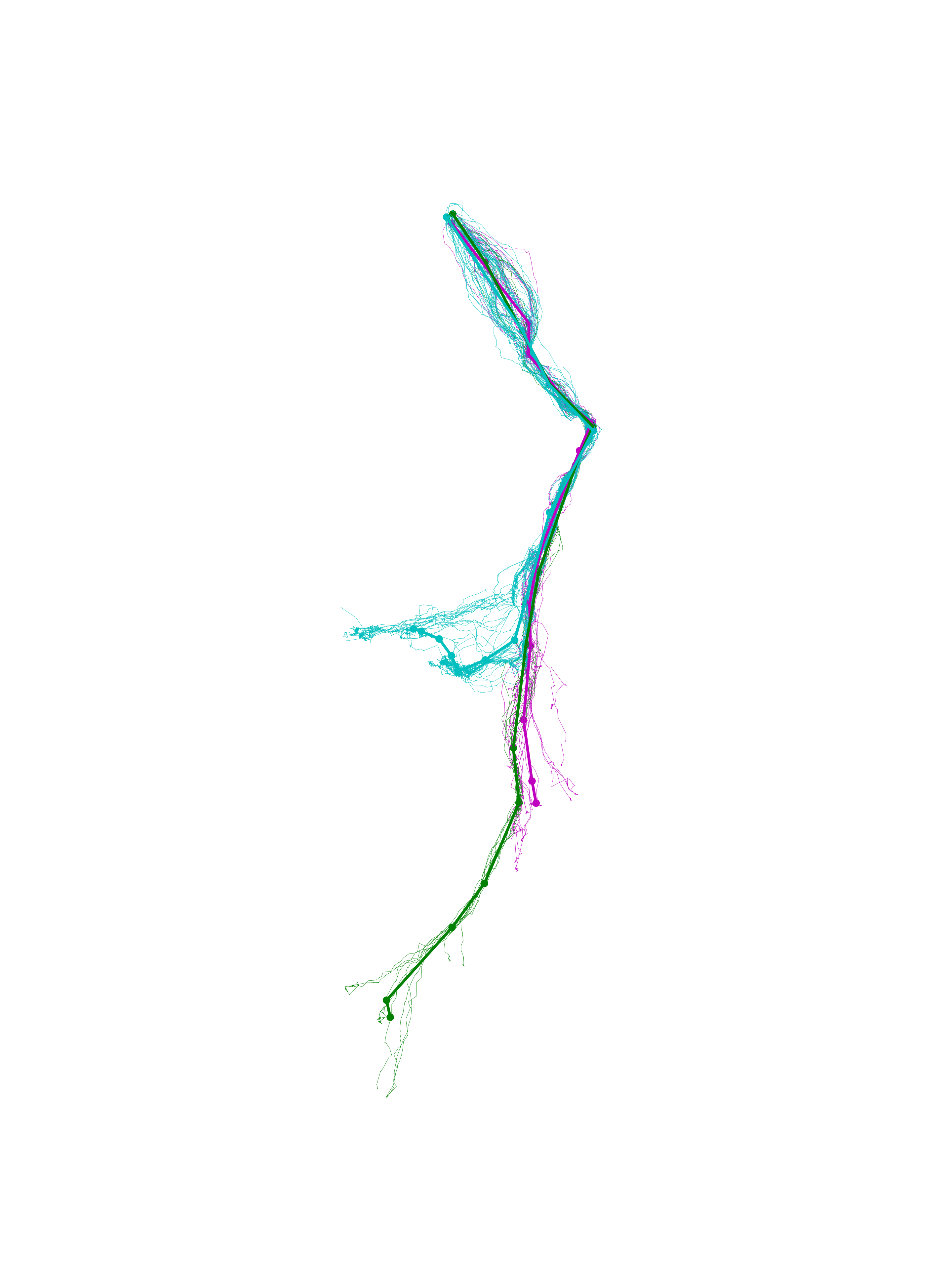}
        \subcaption{Wedge method.}
    \end{minipage}
    \Description{The clusters on the stork migration set.
    The FSA center has noticeable artifacts and does not follow the trajectories in a visually
    convincing way.
    The DBA center ignores a noticeable sideways bend in the trajectories, and so follows
    the trajectories worse visually than CDBA and the wedge method.
    CDBA and the wedge method produce visually very similar results.}
    \caption{Centers of an initial clustering updated with (from left to right) FSA, DBA, CDBA, and
    the wedge method.
    The trajectory data used is the movebank stork migration data set.}
    \label{fig:movebank}
\end{figure}
\end{document}